\documentclass[runningheads]{llncs}

\usepackage{eccv}

\usepackage{eccvabbrv}

\usepackage{graphicx}
\usepackage{booktabs}
\usepackage{bbm}
\usepackage{dsfont}
\usepackage[misc]{ifsym}
\usepackage{url}
\usepackage[accsupp]{axessibility}  % Improves PDF readability for those with disabilities.

\usepackage[table]{xcolor} 
\usepackage{utfsym}
\usepackage{multirow} 
\usepackage{soul, color, xcolor}
\AtBeginDocument{
\setlength\abovedisplayskip{2pt plus 2pt minus 2pt}
\setlength\belowdisplayskip{2pt plus 2pt minus 2pt}
\setlength\abovedisplayshortskip{2pt plus 2pt minus 2pt}
\setlength\belowdisplayshortskip{2pt plus 2pt minus 2pt}
}

\definecolor{mygray}{gray}{.92}
\makeatletter
\def\blfootnote{\xdef\@thefnmark{}\@footnotetext}
\makeatother
\usepackage{hyperref}

\usepackage{orcidlink}

\begin{document}

% ---------------------------------------------------------------
% TODO REVIEW: Replace with your title
\title{ViTexQA: A Multi-Frame Temporal Perception Dataset for Video Text Question Answering}

% TODO REVIEW: If the paper title is too long for the running head, you can set
% an abbreviated paper title here. If not, comment out.
\titlerunning{ViTexQA}

% TODO FINAL: Replace with your author list. 
% Include the authors' OCRID for the camera-ready version, if at all possible.
% \author{First Author\inst{1}\orcidlink{0000-1111-2222-3333} \and
% Second Author\inst{2,3}\orcidlink{1111-2222-3333-4444} \and
% Third Author\inst{3}\orcidlink{2222--3333-4444-5555}}

% % TODO FINAL: Replace with an abbreviated list of authors.
% \authorrunning{F.~Author et al.}
% % First names are abbreviated in the running head.
% % If there are more than two authors, 'et al.' is used.

% % TODO FINAL: Replace with your institution list.
% \institute{Princeton University, Princeton NJ 08544, USA \and
% Springer Heidelberg, Tiergartenstr.~17, 69121 Heidelberg, Germany
% \email{lncs@springer.com}\\
% \url{http://www.springer.com/gp/computer-science/lncs} \and
% ABC Institute, Rupert-Karls-University Heidelberg, Heidelberg, Germany\\
% \email{\{abc,lncs\}@uni-heidelberg.de}}

\author{Zhentao Guo\inst{1}\textsuperscript{$*$}\orcidlink{0009-0000-0242-4059} \and Chen Duan\inst{1}\textsuperscript{$*$}\orcidlink{0000-0003-3346-8315} \and Tongkun Guan\inst{2}\textsuperscript{$*$}\orcidlink{0000-0003-3346-8315} \and
Zining Wang\inst{1}\orcidlink{0009-0007-3847-6396} \and Kai Zhou\inst{1(\textrm{\Letter})}\orcidlink{0009-0002-6051-3791} \and
 Pengfei Yan\inst{1}\orcidlink{0009-0008-3438-8994}
}

% TODO FINAL: Replace with an abbreviated list of authors.
\authorrunning{Z.~Guo et al.}
% First names are abbreviated in the running head.
% If there are more than two authors, 'et al.' is used.

% TODO FINAL: Replace with your institution list.
\institute{Meituan \email{\{guozhentao, duanchen, wangzining, zhoukai, yanpengfei\}@meituan.com} 
\and Shanghai Jiao Tong University \email{gtk0615@sjtu.edu.cn} \\
\url{https://github.com/ZhentaoGuo/ViTexQA}
}

% \maketitle

\maketitle
\blfootnote{\noindent$^{*}$Equal contribution. \textsuperscript{\Letter}Corresponding author.}
\begin{abstract}

Despite remarkable progress in multimodal understanding, current MLLMs still exhibit limitations in video text understanding, particularly when semantics emerge through the integration of temporally distributed textual cues across multiple frames. This perception challenge fundamentally differs from static image text understanding, yet existing datasets fail to capture: the vast majority of questions remain answerable from single frames, inadequately reflecting real-world video text comprehension demands. 
To address this, we present ViTexQA, a large-scale video-text QA dataset, and FrameThinker for robust multi-frame temporal reasoning. We build ViTexQA via a quality-controlled Chain-of-Thought (CoT) annotation pipeline boosted with temporal constraints; all its QA pairs demand cross-frame text fusion to solve, enforcing true temporal reliance.
FrameThinker adopts two-stage training for explicit temporal modeling: CoT-Guided Supervised Fine-Tuning (SFT) generates frame-aware reasoning chains, followed by Temporally-grounded Reinforcement Learning (RL) optimized with multi-frame coherence rewards. Evaluations show our method outperforms SOTA baselines on ViTexQA, lifting ROUGE-L by 6.3\%.

\end{abstract}    
\section{Introduction}
\label{sec:intro}

\begin{figure*}[t]
    \centering
    \includegraphics[width=0.85\linewidth]{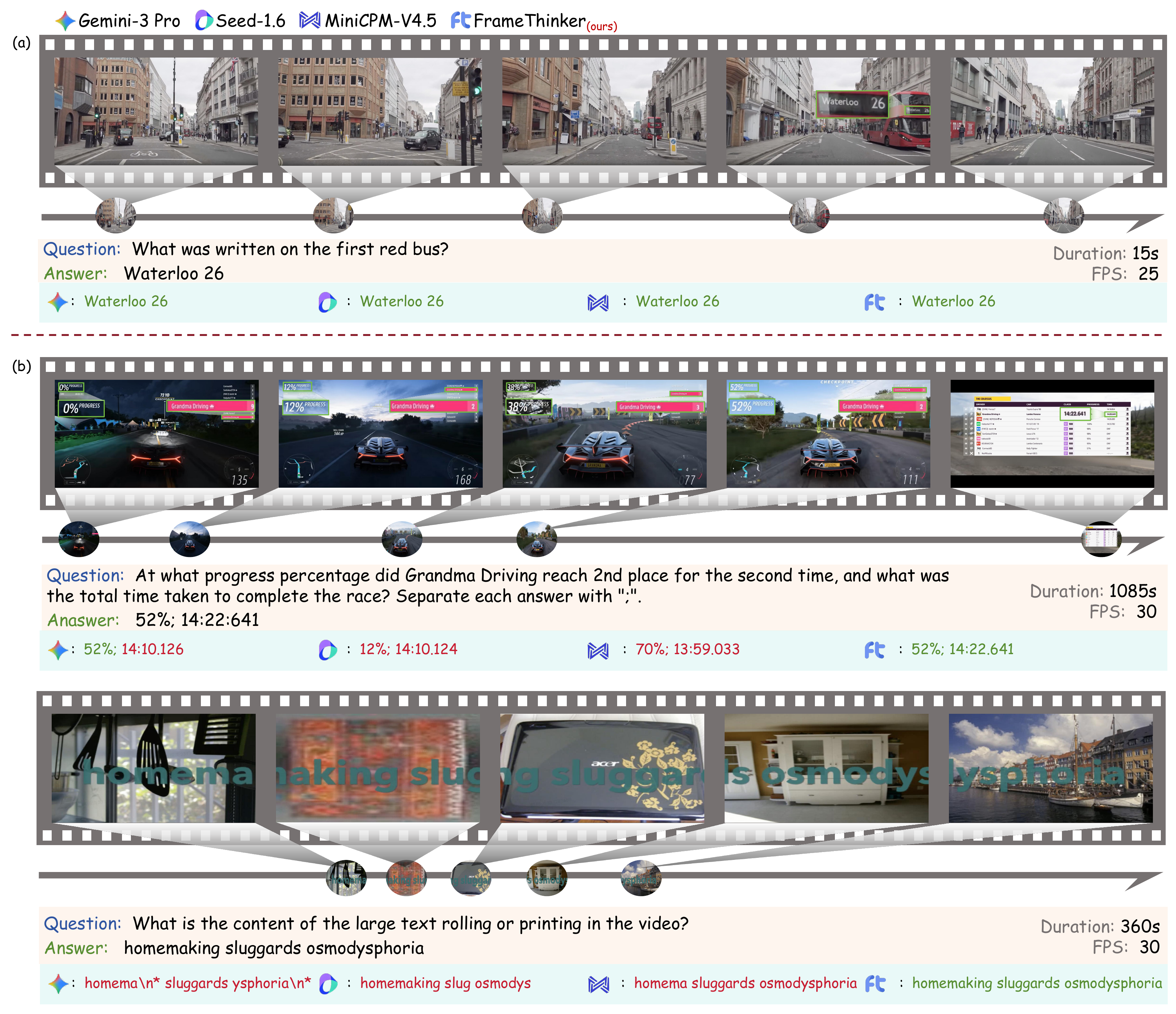}
    \vspace{-1em}
    \caption{Single-frame and multi-frame perception examples. \textbf{(a)} Single-frame perception, where the answer is derived from textual content within a single frame (such as reading ``Waterloo 26'' from the red bus.). All compared models answer correctly on such single-frame perception questions. \textbf{(b)} Multi-frame perception, which requires the model to integrate textual, spatial, and temporal cues across multiple frames (such as finding the answer to the question based on the progress of the competition, ranking changes, and time spent.). In contrast, while other models frequently produce incorrect or incomplete answers, our \textbf{FrameThinker} consistently arrives at the correct answer.}
    \label{fig:firstpage}
    \vspace{-2em}
\end{figure*}
% \vspace{5pt}

% The rapid development of Multimodal Large Language Models (MLLMs)~\cite{bai2025qwen2,chen2024internvl,li2024monkey,liu2023visual} has enabled unified approaches to diverse vision-language tasks, from static image understanding~\cite{singh2019towards,biten2019scene} to dynamic video comprehension~\cite{lei2018tvqa,xu2017video,wang2024internvideo2}. 

The rapid development of Multimodal Large Language Models (MLLMs)~\cite{bai2025qwen2,chen2024internvl,li2024monkey,yao2025efficient,yin2024survey,guan2026codepercept,wang2025marten,guo2026enhanced} has enabled unified approaches to diverse vision-language tasks, from static image understanding~\cite{guan2025token,guan2024posformer,guan2024bridging} to dynamic video comprehension~\cite{xu2017video,wang2024internvideo2,lin2024video,li2024topa,shen2025long,liu2025nvila}.
Among these capabilities, the ability to understand text within visual content has proven essential for real-world applications. While substantial progress has been made in image-based scene text understanding through benchmarks like DocVQA~\cite{mathew2021docvqa}, TextVQA~\cite{singh2019towards}, and ST-VQA~\cite{biten2019scene}, the challenge of comprehending text in videos remains fundamentally different.

Videos introduce a critical temporal dimension that distinguishes them from static images. Textual information in videos often carries semantics that emerge only through temporal integration rather than appearing complete in isolated moments. 
Consider sports broadcasts where understanding the game progression requires tracking score changes across frames, news programs where complete stories unfold through sequential caption updates, or instructional videos where procedural knowledge accumulates through temporally distributed textual cues~\cite{lee2025video,de2025describe}. In these scenarios, comprehension requires synthesizing information dispersed across multiple timestamps to construct complete semantic interpretations. As shown in Fig.~\ref{fig:firstpage}, while single-frame questions can be answered by examining one snapshot, multi-frame perception demands integrating evidence distributed throughout the video timeline to derive meaningful answers.

Despite the importance of temporal perception over textual content, existing video text QA datasets exhibit a critical limitation. Our statistical analysis reveals that only approximately \textbf{13\%} of questions in M4-ViteVQA~\cite{zhao2022towards} and \textbf{18\%} in EgoTextVQA~\cite{zhou2025egotextvqa} genuinely require multi-frame perception. The overwhelming majority can be answered from single frames, failing to capture the temporal complexity inherent in real-world video understanding. This gap reflects a broader issue in current benchmark design: datasets inadvertently allow models to bypass temporal perception by relying on static, frame-level perception. Consequently, MLLMs trained and evaluated on these benchmarks develop limited capabilities for cross-frame integration, hindering their effectiveness in scenarios demanding true temporal comprehension. 

To address this fundamental gap, we introduce ViTexQA, a large-scale dataset explicitly designed to require multi-frame temporal perception over textual content in videos. Unlike existing benchmarks where temporal perception remains optional, ViTexQA ensures that every question is deliberately unanswerable from any single frame alone. Each question necessitates integrating textual cues distributed across multiple timestamps, forcing models to engage in genuine temporal perception rather than frame-level pattern matching. This design philosophy reflects authentic video understanding demands where critical information emerges only through temporal synthesis.

Specifically, ViTexQA is constructed through two complementary pipelines that together address both evaluation and training needs for multi-frame temporal perception. The Video Dataset Construction pipeline ensures genuine multi-frame temporal dependencies through rigorous data collection and quality controlled human annotation, producing 5,147 high-quality videos with 6,864 QA pairs where each question is verified to require cross-frame integration. The Multi-frame perception CoT Construction pipeline further enriches these QA pairs with temporally-grounded CoT annotations that explicitly preserve spatiotemporal textual information across video frames. Together, these pipelines yield a comprehensive dataset spanning approximately 363 hours across diverse domains including sports, news, tutorials, and everyday scenarios. Critically, we adopt a human-centric approach throughout: all annotations are created exclusively by human annotators without automated assistance from GPT or similar systems, ensuring alignment with natural human perception patterns and genuine challenge for current MLLMs~\cite{lin2024video,team2025vidi2,liu2025nvila,shen2025long,li2024topa,guan2025ccdplus}.

Building upon ViTexQA, we further propose FrameThinker, a systematic method to enhance multi-frame temporal perception capabilities in MLLMs. FrameThinker addresses the temporal perception challenge through a two-stage training paradigm. First, CoT-Guided SFT trains models to generate temporally-grounded perception chains that explicitly enumerate textual evidence across frames with spatial and temporal grounding. Second, Temporally-grounded RL refines these capabilities through composite reward mechanisms that encourage temporal coherence, multi-frame dependency, and accurate evidence aggregation. This approach enables models to develop explicit temporal perception strategies rather than relying on implicit frame-level correlations.

Finally, we conduct extensive experiments evaluating state-of-the-art MLLMs on ViTexQA. Results reveal substantial performance gaps compared to human accuracy, highlighting fundamental limitations in current models' temporal perception abilities. Furthermore, FrameThinker achieves significant improvements over strong baselines, demonstrating the effectiveness of explicit temporal perception training. These findings underscore both the challenge posed by genuine multi-frame temporal perception and the potential for systematic capability enhancement through appropriate training methodologies.

Our contributions are summarized as follows:
\begin{itemize}
    % \item We present ViTexQA, the first large-scale dataset achieving 100\% verified multi-frame dependency, constructed through dual complementary pipelines producing both rich question-answer pairs and corresponding temporally-grounded CoT annotations via rigorous human-centric annotation.
    \item We present ViTexQA, the first large-scale dataset tailored for multi-frame temporal awareness,  constructed through dual complementary pipelines producing both rich question-answer pairs and corresponding temporally-grounded CoT annotations via rigorous human-centric annotation.
    \item We propose FrameThinker, a training methodology that systematically enhances multi-frame temporal perception through CoT-guided SFT and Temp- orally-grounded RL.
    \item Comprehensive benchmarking on ViTexQA reveals critical gaps in current MLLMs' temporal perception capabilities. In contrast, extensive experiments demonstrate that FrameThinker effectively mitigates these limitations, significantly outperforming strong baselines and validating the effectiveness of explicit temporal perception training for video text understanding.
\end{itemize}

\section{Related Work}
\label{sec:formatting}

%-------------------------------------------------------------------------
\noindent \textbf{Image-based text VQA datasets}
OCR-related datasets are among the most important benchmarks for evaluating MLLMs in visual text understanding. In the OCR domain, a variety of datasets have been proposed to evaluate different aspects of visual text understanding. Natural scene text VQA datasets, such as TextVQA~\cite{singh2019towards}, ST-VQA~\cite{biten2019scene}, EST-VQA~\cite{wang2020general}, and OCR-VQA~\cite{mishra2019ocr}, focus on question answering over images containing text in natural scenes, where the model must integrate scene understanding with optical character recognition. Document and chart understanding tasks address more structured and information-dense data types. For example, InfoVQA~\cite{mathew2022infographicvqa} requires models to jointly reason over  textual content, graphical elements, and data visualizations; ChartQA~\cite{masry2022chartqa} is designed for question answering over charts, demanding numerical perception and chart-specific knowledge; DocVQA~\cite{mathew2021docvqa} focuses on document-level text understanding and retrieval-based question answering. Web-based reading comprehension datasets, such as WebSRC~\cite{chen2021websrc}, target the understanding of web page layouts and interactive elements. In addition, several comprehensive OCR benchmarks have been introduced, such as OCRBench~\cite{liu2024ocrbench}, OCRBench v2~\cite{fu2024ocrbench}, and SEED-Bench-2-Plus~\cite{li2024seed}, which integrate multiple OCR-related tasks to provide a holistic evaluation of MLLMs' text understanding capabilities~\cite{hu2025mplug,alampara2025probing}.

%-------------------------------------------------------------------------
\noindent \textbf{Video-based text VQA datasets} While image-based Text VQA has been extensively studied, real-world scenarios often involve dynamic visual content where textual information appears across multiple frames. In videos, text serves as a crucial semantic carrier, complementing object and action cues with information that is otherwise visually inaccessible~\cite{zhao2022towards,nguyen2014video,lee2025video,de2025describe,ozaki2025bqa}. Benchmarking efforts in video text understanding are diverse and comprehensive. For example, MME-VideoOCR~\cite{shi2025mme}, VidText~\cite{yang2025vidtext}, FG Bench~\cite{fei2025current}, and EgoTextVQA~\cite{zhou2025egotextvqa} provide comprehensive evaluation suites that assess model performance from multiple perspectives, including text recognition, visual text question answering, and text grounding~\cite{wang2024internvideo2,nguyen2025horus,ozaki2025bqa}. These benchmarks offer valuable insights into the strengths and weaknesses of existing approaches. In contrast, datasets specifically designed for video text question answering remain scarce. RoadTextVQA~\cite{tom2023reading} focuses on driver assistance scenarios, where questions are answered based on textual information such as road signs present in driving videos. NewsVideoQA~\cite{jahagirdar2023watching} introduces text-based question answering in news videos, requiring systems to analyze embedded textual content. M4-ViteVQA~\cite{zhao2022towards} comprises 7,620 video clips across nine scenes. However, only \textbf{13\%} to \textbf{18\%} of the questions in these datasets require the integration of information from multiple frames, which limits their capacity to evaluate temporal perception abilities. Recognizing the importance of multi-frame context for text-rich video question answering, we introduce ViTexQA, a dataset purpose-built for assessing MLLMs’ temporal textual perception.

\begin{figure*}[t]
    \centering
    \includegraphics[width=0.99\linewidth]{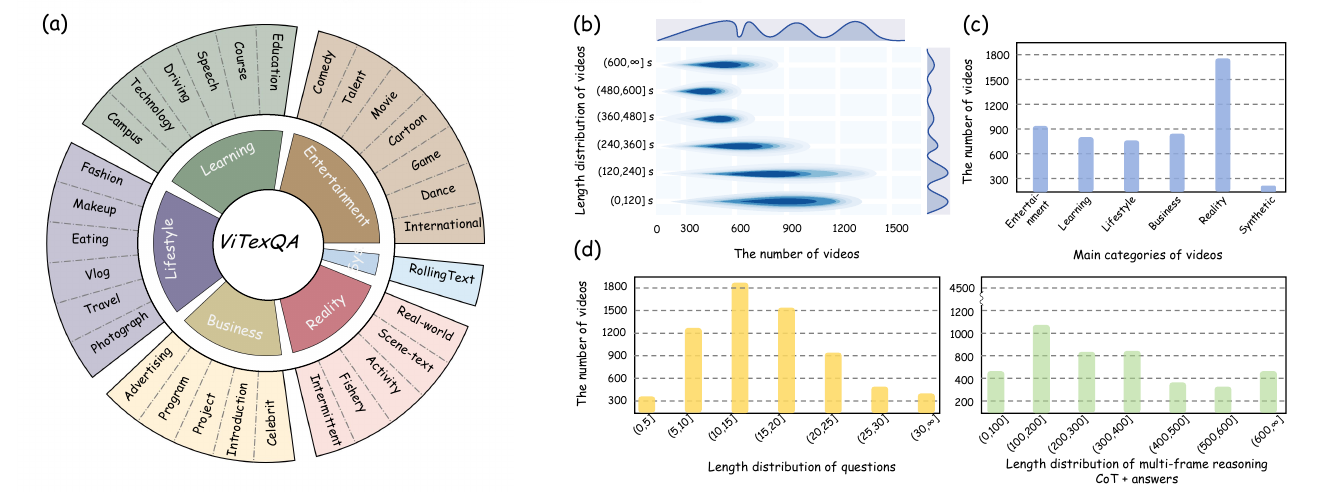}
    \caption{Statistical overview of our ViTexQA dataset. \textbf{(a)} Video genre taxonomy with six major categories and their subcategories. \textbf{(b)} Distribution of video durations. \textbf{(c)} Video count distribution across six major categories. \textbf{(d)} Length distributions of questions \textbf{(left)} and CoT annotations with answers \textbf{(right)} measured in word count.}
    \label{fig:Overview}
    \vspace{-2em}
\end{figure*}

\section{ViTexQA Dataset}
The ViTexQA dataset is constructed through two complementary pipelines: 

\noindent \textbf{(1) Video Dataset Construction pipeline} that ensures genuine multi-frame temporal perception requirements through rigorous data collection and quality-controlled annotation, producing 5,147 high-quality videos and 6,864 QA pairs with verified temporal dependencies;

\noindent \textbf{(2) Multi-frame perception CoT Construction pipeline} that further yields temporally-grounded Chain-of-Thought (CoT) annotations by preserving spatiotemporal text information across video frames to enrich the perception chain of provided QA pairs. 

Together, these pipelines produce a comprehensive dataset (ViTexQA) with diverse video sources, enforced multi-frame dependencies, and temporally-grounded perception CoT annotations, as illustrated in Fig.~\ref{fig:Overview}. Compared to existing datasets or benchmarks, ViTexQA uniquely focuses on multi-frame temporal perception, requiring models to integrate text information across multiple video frames rather than relying on single-frame understanding, a critical capability where current MLLMs exhibit significant weaknesses. As shown in Table~\ref{tab:benchmark}, while prior benchmarks offer limited or no multi-frame perception requirements, ViTexQA ensures multi-frame dependency through rigorous verification protocols, providing both evaluation data and rich CoT training annotations (6,864 QA pairs from 5,147 videos) to address this fundamental gap in video text understanding~\cite{yin2024survey,alampara2025probing}.

\begin{table*}[t]
\centering
\caption{Datasets comparison for video text understanding. }
\label{tab:benchmark}
\scalebox{0.85}{ 
\begin{tabular}{ccccc}  % 改用tabular
\hline
\textbf{Dataset} & \textbf{\#Scenarios} &\textbf{\#Videos}& \textbf{\#QA}   & \textbf{Designed for multi-frame perception?}  \\
\hline
\multicolumn{5}{c}{Train Datasets} \\
\hline
% DSText V2~\cite{wu2024dstext}  & 7   & 90 & -   &  -    & -  \\
% BovText~\cite{wu2021bilingual}        & 32  & 1,541  & -       &    -      & -  \\
M4-ViteVQA~\cite{zhao2022towards}     & 9   & 5,444  & 18,200    & No \\
RoadTextVQA~\cite{tom2023reading}     & 1   & 2,557  & 8,393     &  No  \\
NewsVideoQA~\cite{jahagirdar2023watching}  & 1   & 2,407 & 6,994   & No \\
\hline
 \rowcolor[HTML]{E5F4FB}
\textbf{ViTexQA}                      & 30  & 4,472  & 6,124     &  Yes \\
\hline
\multicolumn{5}{c}{Evaluation Benchmarks} \\
\hline
OCR Benchmark~\cite{fu2024ocrbench}   & 20+ & 25   & 1,477     & No \\
MME-VideoOCR~\cite{shi2025mme}        & 44  & 1,464 & 2,000  & No \\
% BovText~\cite{wu2021bilingual}        & 32  & 480  & -       &    -      & -  \\
% RoadText-1K~\cite{reddy2020roadtext}  & 1   & 1,000 & -       &   -   & -  \\
M4-ViteVQA~\cite{zhao2022towards}     & 9   & 680  & 2,103     & No \\
RoadTextVQA~\cite{tom2023reading}     & 1   & 329  & 1,052     & No \\
EgoTextVQA~\cite{zhou2025egotextvqa}  & 2   & 1,507 & 7,064     & No \\
NewsVideoQA~\cite{jahagirdar2023watching}  & 1   & 346 & 964   & No \\
VidText~\cite{yang2025vidtext}        & 27  & 939  & 2,857    & No \\
\hline
 \rowcolor[HTML]{E5F4FB}
\textbf{ViTexQA}                      & 30  & 675  & 740     & Yes \\
\hline
\end{tabular}
}
\vspace{-1em}
\end{table*}

\begin{figure*}[t]
    \centering
    \includegraphics[width=0.95\linewidth]{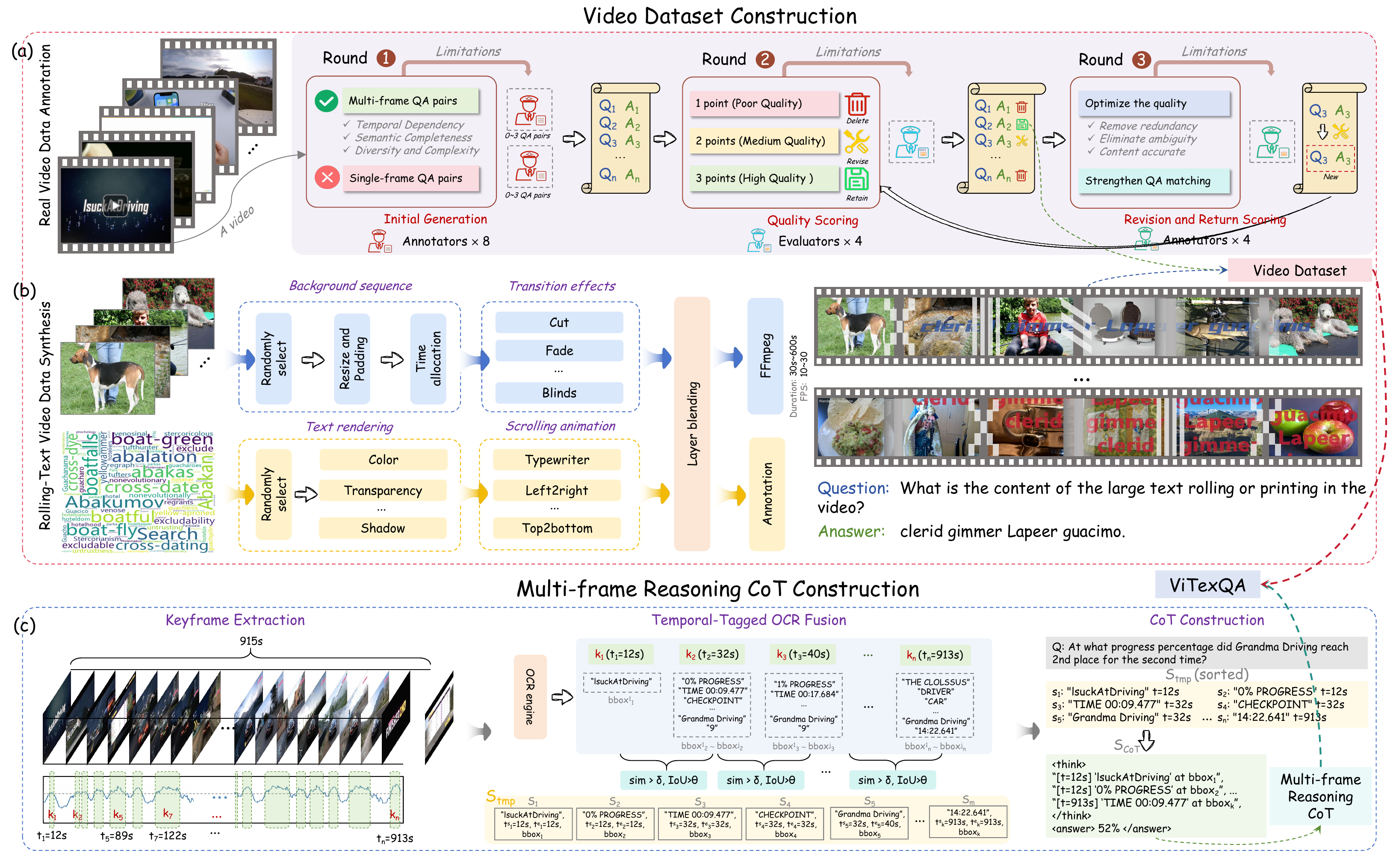}
    \caption{Video Dataset Construction pipeline of our proposed ViTexQA. (a) Manual three-round annotation pipeline for real videos. (Round 1)  Initial Generation. (Round 2) Quality Scoring. (Round 3) Revision and Return Scoring. (b) Synthetic scrolling text video pipeline. By randomly selecting background images and text corpus, setting random rendering parameters and scrolling methods, it generates 30s$\sim$600s random videos with QA annotations. (c) Multi-frame perception CoT construction pipeline. The text information, timestamps, and location annotations of multi-frame are integrated into $\mathrm{S}_{\rm cot}$ through three steps.}
    \label{fig:Pipeline_sys}
    \vspace{-2em}
\end{figure*}

\subsection{Video Dataset Construction}
To construct a high-quality benchmark for genuine multi-frame temporal text perception, we establish a rigorous data curation and annotation framework encompassing diverse video sources, quality-controlled annotation protocols, and strategic synthetic augmentation.

\noindent \textbf{(1) Data Collection and Curation.} Videos are sourced from: (i) YouTube across 30 diverse domains (games, fashion, technology, etc.), selected for rich temporal text patterns~\cite{lee2025video,lian2023llm}; (ii) challenging cases from existing datasets with original annotations discarded; (iii) To enhance temporal text pattern coverage, we developed a Rolling-Text synthesis pipeline (Fig.~\ref{fig:Pipeline_sys}(b)) that generates 100 synthetic scrolling text videos.
All videos were standardized to 720p resolution to ensure text clarity and uniform quality. The dataset is strictly for research purposes and complies with data privacy policies.

\noindent \textbf{(2) Quality-Controlled Annotation Pipeline.} 
As shown in Fig.~\ref{fig:Pipeline_sys}(a), real videos undergo iterative three-round annotation. (Round 1) Eight trained annotators independently create question-answer pairs with mandatory multi-frame perception requirements. Each video receives annotations from two different annotators to capture diverse perception perspectives. Annotators are explicitly instructed to verify that questions are unanswerable from any single frame alone, enforced through frame-by-frame answerability checks. (Round 2) Four senior evaluators assess each annotation on three criteria—temporal dependency necessity (does the question require cross-frame integration?), answer correctness, and question clarity—assigning scores: 1 (reject: single-frame answerable or low quality), 2 (revision needed: minor issues in temporal grounding or clarity), 3 (accept: meets all quality standards). (Round 3) Score-2 annotations undergo targeted revision by four expert annotators, then return to Round 2 for re-evaluation. This iterative process continues until all samples achieve score-3, ensuring dataset-wide quality consistency.

\subsection{Multi-frame Perception CoT Construction}
Building upon the quality-controlled Video QA pairs described above, we propose a CoT data generation pipeline to enable multi-frame temporal perception. This pipeline preserves spatiotemporal text information across frames, providing detailed perception annotations for each question-answer pair in Fig.~\ref{fig:Pipeline_sys}(c).

% \subsubsection{Keyframe extraction.}
% Given a video sequence of duration $T$ seconds, we aim to extract a sparse set of keyframes $\mathcal{K} = \{k_1, k_2, \ldots, k_n\}$ to reduce computational overhead while preserving salient temporal information. We employ a hash-based detection~\cite{sadowski2007simhash} method to identify frames with significant visual changes. Specifically, for consecutive frames $f_i$ and $f_{i+1}$, we compute the perceptual hash difference:
% \begin{equation}
% \Delta H_i = d_H(H(f_i), H(f_{i+1}))
% \end{equation}
% where $d_H$ denotes the Hamming distance metric. $H(\cdot)$ denotes the perceptual hash function. Frames are selected as keyframes when $\Delta H_i$ exceeds a predefined threshold $\tau$, ensuring that only frames with substantial visual content changes are retained. This approach effectively reduces redundancy while maintaining temporal coherence across the video.

\noindent \textbf{Keyframe extraction.}
Given a video sequence of duration $T$ seconds, we aim to extract a sparse set of keyframes $\mathcal{K} = \{\mathbf{k}_1, \mathbf{k}_2, \ldots, \mathbf{k}_n\}$ to reduce computational overhead while preserving salient temporal information. We employ a hash-based detection~\cite{sadowski2007simhash} to identify frames with significant visual changes. Specifically, for consecutive frames $\mathbf{f}_i$ and $\mathbf{f}_{i+1}$, we compute the perceptual hash difference:
\begin{equation}
\Delta H_i = \mathrm{d_H}(\mathrm{H}(\mathbf{f}_i), \mathrm{H}(\mathbf{f}_{i+1}))
\end{equation}
where $\mathrm{d_H}$ denotes the Hamming distance metric and $\mathrm{H}(\cdot)$ denotes the perceptual hash function. Frames are selected as keyframes when $\Delta H_i$ exceeds a predefined threshold $\tau$, ensuring that only frames with substantial visual content changes are retained. This approach effectively reduces redundancy while maintaining temporal coherence across the video.

\noindent \textbf{Temporal-Tagged OCR Fusion.}
Conventional OCR fusion methods~\cite{zhao2022towards} typically perform simple text deduplication, discarding temporal information about ``when'' and ``where'' text appears. However, in multi-frame perception scenarios, the temporal order and spatial location of text are crucial for correct inference. We propose Temporal-Tagged OCR Fusion to explicitly preserve spatiotemporal metadata for each detected text instance.

For each keyframe $\mathbf{k}$ (omit the subscript for notation simplicity) of all extracted keyframe $\mathcal{K} = \{\mathbf{k}_1, \mathbf{k}_2, \ldots, \mathbf{k}_n\}$, we apply an OCR engine~\cite{du2020pp} to detect text regions. The $j$-th detected OCR text instance in keyframe $\mathbf{k}$ is defined:
\begin{equation}
\mathbf{o}_j = (\mathtt{text}_j, t, \mathtt{bbox}_j)
\end{equation}
where $\mathtt{text}_j$ is the recognized text, $t$ is the corresponding timestamp, $\mathtt{bbox}_j$ is the bounding box. The complete OCR detection set is $\mathcal{O}_{\mathbf{k}} = \{\mathbf{o}_j \mid \mathbf{k} \in \mathcal{K},\ j = 1, 2, \ldots, n\}$.

% Then we define two text instances $\mathbf{o}_j$ and $\mathbf{o}_p$ as temporally equivalent if:
Then, we define two text instances $\mathbf{o}_p$ and $\mathbf{o}_q$ from different keyframes as content-equivalent if the following condition is met:
\begin{equation}
\mathrm{sim}(\mathtt{text}_p, \mathtt{text}_q) > \delta \quad \text{and} \quad \mathrm{IoU}(\mathtt{bbox}_p, \mathtt{bbox}_q) > \theta
\end{equation}
where $\mathrm{sim}(\cdot, \cdot)$ measures string similarity and $\mathrm{IoU}(\cdot, \cdot)$ computes bounding box overlap. Based on this equivalence relation, all text instances in $\mathcal{O}_{\mathbf{k}}$ are grouped into $M$ content-equivalent classes $\{\mathcal{C}_1, \mathcal{C}_2, \ldots, \mathcal{C}_M\}$, where each class $\mathcal{C}_m$ ($m = 1, 2, \ldots, M$) contains text instances that are mutually content-equivalent across multiple keyframes. For each class $\mathcal{C}_m$, we create a time text span:
\begin{equation}
\mathbf{s}_m = (\mathtt{text}_m,\ t_m^{\mathrm{s}},\ t_m^{\mathrm{e}},\ \mathtt{bbox}_m)
\end{equation}
where $t_m^{\mathrm{s}} = \min_{\mathbf{o} \in \mathcal{C}_m} t(\mathbf{o})$ and $t_m^{\mathrm{e}} = \max_{\mathbf{o} \in \mathcal{C}_m} t(\mathbf{o})$ denote the first and last appearance times. Then, sort the data by time according to $t_m^{\mathrm{s}}$ to obtain the time corpus $\mathcal{S}_{\mathrm{tmp}}= \{\mathbf{s}_1, \mathbf{s}_2, \ldots, \mathbf{s}_M\}$. This representation retains both the temporal sequence and persistence of text, which are critical for multi-frame perception.

\noindent \textbf{CoT Construction.}
Given a question $Q$ with ground-truth answer $A$ and the temporal corpus $\mathcal{S}_{\mathrm{tmp}}$ obtained from the previous fusion step, we construct perception CoT annotations that explicitly capture the multi-frame temporal perception process. Specifically, for each temporal text span $\mathbf{s}_m = (\mathtt{text}_m, t_m^{\mathrm{s}}, \\t_m^{\mathrm{e}}, \mathtt{bbox}_m)$ in $\mathcal{S}_{\mathrm{tmp}}$, we construct a structured perception trace that interleaves temporal information with textual content. The complete CoT sequence is formulated as:
\begin{equation}
\langle \texttt{<think>} \; \bigoplus_{x=1}^{N} \texttt{``[t=}t_x^{\mathrm{s}}\texttt{s] `}\mathtt{text}_{\mathrm{gt}}^x\texttt{' at }\mathtt{bbox}_{\mathrm{gt}}^x\texttt{''} \; \texttt{</think>} \rangle
\end{equation}

where $\bigoplus$ denotes temporal concatenation ordered by appearance time $t_x^{\mathrm{s}}$. $\mathtt{text}_{\mathrm{gt}}^x$ and $\mathtt{bbox}_{\mathrm{gt}}^x$ represent the recognized text and bounding box of the $t_x^{\mathrm{s}}$ timestamp, respectively.

% This temporally-tagged format serves three critical purposes. First, it explicitly encodes ``when'' each piece of text information appears through timestamps $t_k^{\mathrm{s}}$, enabling models to understand temporal ordering and dependencies. Second, it specifies ``where'' the text is located via bounding boxes $\mathtt{bbox}_k$, grounding visual attention to relevant spatial regions. Third, it preserves ``what'' textual content ($\mathtt{text}_k$) is present at each moment, maintaining the complete evidence chain needed for answer derivation.

To ensure perception quality and temporal coherence, we then connect the temporal evidence to the final answer. The resulting temporally-grounded CoT annotations provide rich supervision signals that teach models not only what to predict, but how to integrate text information across multiple frames through explicit temporal perception. This representation fundamentally differs from existing single-frame text understanding approaches, as it requires models to track information flow across time and reason about temporal dependencies between text instances.
\section{FrameThinker}
\begin{figure*}[t]
    \centering
    \includegraphics[width=0.95\linewidth]{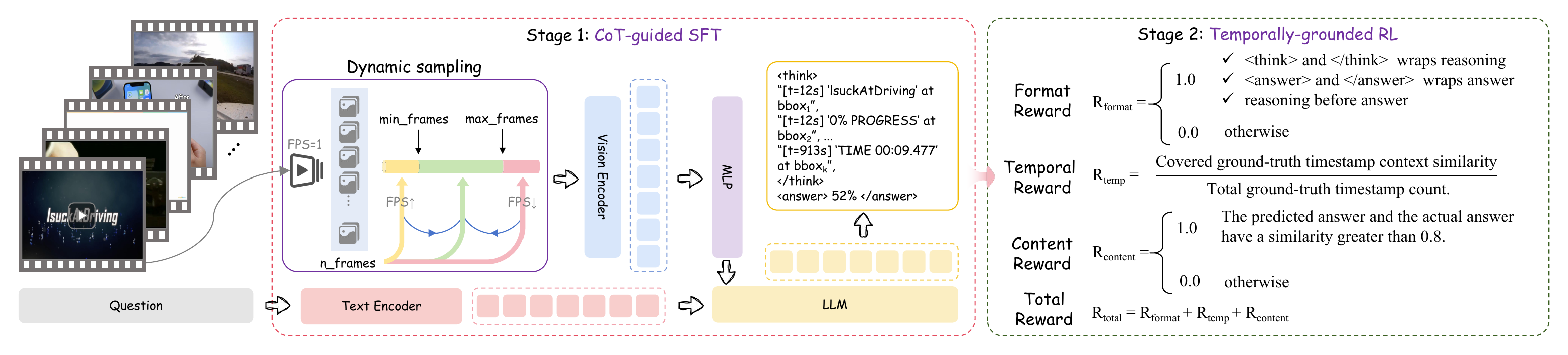}
    \caption{Overall pipeline of FrameThinker. In stage 1, video frames sampled via dynamic sampling are fed into multi-frame perception CoT-guided SFT. In stage 2, following the standard GRPO approach, $r_{\mathrm{fmt}}$, $r_{\mathrm{tmp}}$, and $r_{\mathrm{cnt}}$ are designed to enable the model to be further optimized for multi-frame perception.}
    \label{fig:Pipeline}
    \vspace{-1em}
\end{figure*}

Build upon ViTexQA, we further propose FrameThinker, a multi-modal training method for multi-frame video text understanding. As illustrated in Fig.~\ref{fig:Pipeline}, FrameThinker takes uniformly sampled video frames and temporal-aware CoT annotations as input, and employs a two-stage training paradigm: CoT-guided Supervised Fine-Tuning (SFT) for perception consistency, followed by Temporally-grounded Reinforcement Learning (RL) for task-specific optimization. By grounding predictions in explicit spatiotemporal textual evidence, FrameThinker achieves improved accuracy and interpretability for complex video understanding tasks.

% Panel (a) shows the Chain-of-Thought Data Generation pipeline. Through Keyframe extraction and Temporal-Tagged OCR Fusion, it constructs CoT with temporal text information. Panel (b) shows the 

\subsection{CoT-guided SFT}
In the first stage, we perform SFT on video-text pairs augmented with our generated temporal CoT annotations. The model input consists of two modalities: (1) visual information from uniformly sampled video frames $\mathcal{F} = \{\mathbf{f}_1, \mathbf{f}_2, \ldots, \mathbf{f}_N\}$, and (2) textual information comprising the question $Q$ and the temporal-aware CoT constructed from $\mathcal{S}_{\mathrm{tmp}}$. Following standard practices in multimodal learning~\cite{liu2023visual}, visual features are first projected through a multilayer perceptron (MLP), then concatenated with text token embeddings, and finally fed into the language model backbone for autoregressive generation. Specifically, our training objective is to maximize the likelihood of generating the complete perception-answer sequences:
\begin{align}
    \mathcal{L}_{\mathrm{SFT}} = -\mathbb{E}_{(Q, \mathcal{F}, \mathrm{S}_{\rm cot}) \sim \mathcal{D}} \left[ \log \pi_{\theta}\left(\mathrm{S}_{\rm cot} \mid Q, \mathcal{F}\right) \right]
\end{align}
where $\mathrm{S}_{\rm cot} = \big\langle \texttt{<think>} \; \bigoplus_{x=1}^{N} \texttt{``[t=}t_x^{\mathrm{s}}\texttt{s] `}\mathtt{text}_{\mathrm{gt}}^x\texttt{' at }\mathtt{bbox}_{\mathrm{gt}}^x\texttt{''} \; \texttt{</think>} \; 
    \texttt{<answer>}\\ \; \mathrm{S}_{\rm gt}  \; \texttt{</answer>} \big\rangle$.
$\theta$ denotes model parameters. This formulation encourages the model to explicitly articulate its temporal perception process by: (1) enumerating all detected text with spatiotemporal tags, (2) identifying relevant temporal patterns such as sequential order and co-occurrence, and (3) deriving the answer grounded in multi-frame evidence. Unlike conventional approaches that directly predict answers from visual features, CoT-guided SFT enforces intermediate perception steps that mirror human cognitive processes, thereby improving both accuracy and interpretability.

\subsection{Temporally-grounded RL}

In the second stage, we apply Temporally-grounded RL via GRPO~\cite{liu2024deepseek} to further refine the model beyond the limitations of supervised imitation. While CoT-guided SFT provides a strong foundation by teaching the model to mimic human-annotated perception patterns, it is inherently constrained by the quality and coverage of training annotations. RL addresses these limitations by enabling the model to explore alternative perception paths and optimize multi-frame perception through reward-based feedback.

Specifically, we design three complementary reward functions targeting different aspects of output quality, each addressing a specific dimension of the desired model behavior:

\noindent \textbf{Format Reward ($r_{\mathrm{fmt}}$).}
The output format comprises two essential components:
(i) final output format, typically encapsulated within \texttt{<answer>} \texttt{</answer>} tags,
and (ii) the intermediate perception process format, typically enclosed within
\texttt{<think>} \texttt{</think>} tags. Furthermore, the proper nesting order is enforced
such that the perception section strictly precedes the answer section. Formally, the
format reward is defined as:
\begin{equation}
r_{\mathrm{fmt}}=
\begin{cases}
1,  & \text{if format is correct},\\
0, & \text{if format is incorrect}.
\end{cases}
\label{eq:format_reward}
\end{equation}

\noindent \textbf{Temporal Reward ($r_{\mathrm{tmp}}$).}
The temporal grounding reward quantifies how well the model's perception chain references specific temporal evidence from the video, while ensuring that each referenced timestamp is semantically consistent with its surrounding context in the predicted CoT. 

Assump the model output $\mathrm{P}_{\rm cot} = \big\langle \texttt{<think>} \; \bigoplus_{x=1}^{N} \texttt{``[t=}t_x^{\mathrm{s}}\texttt{s] `}\mathtt{text}_{\mathrm{pre}}^x\texttt{' at }\\\mathtt{bbox}_{\mathrm{pre}}^x\texttt{''} \; \texttt{</think>} \; 
    \texttt{<answer>} \; \mathrm{S}_{\rm pre}  \; \texttt{</answer>} \big\rangle$,where $\mathtt{text}_{\mathrm{pre}}^x$ and $\mathtt{bbox}_{\mathrm{pre}}^x$ represent the predicted text and predicted bounding box of the $t_x^{\mathrm{s}}$ timestamp, respectively. Then the temporal reward is defined as:
\begin{equation}
    r_{\mathrm{tmp}} = \frac{1}{|\mathcal{T}_{\rm gt}|} \sum_{t \in \mathcal{T}_{\rm gt}} 
    \mathbb{I}[t \in \mathcal{T}_{\rm pre}] \cdot 
    \mathrm{sim}(\texttt{text}_{\mathrm{pre}}, \texttt{text}_{\mathrm{gt}})
\end{equation}
where $\mathcal{T}_{\mathrm{pre}}$ denotes the set of predicted timestamps in $\mathrm{P}_{\mathrm{cot}}$, $\mathcal{T}_{\mathrm{gt}}$ denotes the set of ground-truth timestamps in $\mathrm{S}_{\mathrm{cot}}$. $\texttt{text}_{\mathrm{pre}}$ and $\texttt{text}_{\mathrm{gt}}$ denote the textual context surrounding timestamp $t$ in the predicted and ground-truth text, respectively.

% , and $\mathrm{sim}(\cdot, \cdot)$ is a similarity function. 

\noindent \textbf{Content Reward ($r_{\mathrm{cnt}}$).}
The content reward evaluates the semantic correctness of the final answer:
\begin{equation}
    r_{\mathrm{cnt}} = \mathbb{I}[\mathrm{sim}(\mathrm{S}_{\rm pre}, \mathrm{S}_{\rm gt}) > \gamma]
\end{equation}
where $\mathrm{S}_{\rm pre}$ is the predicted answer, $\mathrm{S}_{\rm gt}$ is the ground-truth answer.

The threshold $\gamma$ is set empirically (typically $\gamma = 0.8$) to balance strictness and robustness. This binary reward ($r_{\mathrm{cnt}} \in \{0, 1\}$) provides a clear training signal for answer correctness while allowing reasonable flexibility in answer formulation.

\noindent \textbf{Overall Reward and Optimization}
The overall reward is therefore defined as: $r = r_{\mathrm{fmt}} + r_{\mathrm{tmp}} + r_{\mathrm{cnt}}$.

Consider the standard GRPO approach, it samples a group of generated output sequences $\{z_1, z_2, \ldots, z_G\}$ for each multimodal input query $(Q, \mathcal{F}) \in \mathcal{D}$ from the old policy model $\pi_{\theta_{\mathrm{old}}}$. Then GRPO maximizes the following objective to optimize the model $\pi_{\theta}$:
\begin{multline}
    \mathcal{J}(\theta) = \mathbb{E}_{\{z_i\}_{i=1}^G \sim \pi_{\theta_{\mathrm{old}}}(\cdot \mid Q, \mathcal{F})} 
    \Bigg[ \frac{1}{G} \sum_{i=1}^G \Bigg( 
    \min \left( \frac{\pi_\theta(z_i \mid Q, \mathcal{F})}{\pi_{\theta_{\mathrm{old}}}(z_i \mid Q, \mathcal{F})} \mathcal{A}_i, \right.\\
    \left. \mathrm{clip} \left( \frac{\pi_\theta(z_i \mid Q, \mathcal{F})}{\pi_{\theta_{\mathrm{old}}}(z_i \mid Q, \mathcal{F})}, 1-\epsilon, 1+\epsilon \right) \mathcal{A}_i \right)
    - \beta\, \mathbb{D}_{\mathrm{KL}}(\pi_\theta \parallel \pi_{\mathrm{ref}}) \Bigg) \Bigg]
\end{multline}
where $\epsilon$ and $\beta$ are the PPO clipping hyper-parameter and the coefficient controlling the Kullback-Leibler (KL) penalty, respectively. $\pi_{\mathrm{ref}}$ is the reference model initialized from the SFT stage. To avoid notation conflict with the ground-truth answer $A$, we denote the advantage as $\mathcal{A}_i$. Specifically, for a group of $\mathcal{G}$ outputs $\{z_1, \dots, z_G\}$ sampled from the same input $(Q, \mathcal{F})$, the advantage is calculated:
\begin{equation}
    \mathcal{A}_i = \frac{r^{(i)} - \mathrm{mean}(r^{(1)}, r^{(2)}, \dots, r^{(G)})}{\mathrm{std}(r^{(1)}, r^{(2)}, \dots, r^{(G)})}
\end{equation}
where $r^{(i)}$ represents the total reward evaluated on the $i$-th generated output $z_i$. Through this RL stage, the model is encouraged to generate not only structurally compliant CoT perception paths, but also temporally grounded and semantically accurate answers that faithfully interpret the complex multi-frame content.
\section{Experiments}

% 审稿人会不会问 framethinker是用的CoT推理的，其他的是不是用的也是cot。

% We conduct a comprehensive evaluation of 15 large multimodal models using our ViTexQA test set, encompassing both open-source and proprietary models. For closed-source models, we evaluated them using the official multi-image evaluation API. For open-source models, we select current state-of-the-art video LMMs with diverse architectures and LLM sizes, enabling a broad assessment of video text understanding capabilities. All evaluations are conducted in a zero-shot manner.  

% Additionally, we fine-tuned a subset of representative open-source models on our ViTexQA training set and evaluated their performance on the ViTexQA test set and other benchmarks to  assess the effectiveness of task-specific training. More details about the evaluation settings are provided in the Supplementary Materials.

% Finally, we conducted ablation studies to demonstrate the effectiveness of each module of FrameThinker.

\noindent \textbf{Implementation details.}
We use the latest Qwen3-VL~\cite{qwen3} as the base model. All experiments were performed on eight NVIDIA A100 GPUs with 80GB of memory each. Additional details about dataset curation, training settings, and visualizations are provided in supplementary materials.
% FrameThinker apply SigLIP-2~\cite{tschannen2025siglip} and Qwen3 backbone~\cite{yang2025qwen3} by default. Following Qwen2.5-VL~\cite{bai2025qwen25vltechnicalreport}, we deploy a two-layer MLP as the text encoder (visual-language merger) using the pre-trained Qwen3 text encoding module~\cite{yang2025qwen3}. 2×2 visual features from the visual encoder are compressed into single visual tokens by the MLP to align with the hidden dimension of the LLM. We use the latest Qwen3-VL~\cite{qwen3} as the base model. The model was trained with bfloat16 precision and utilized Flash Attention 2 for efficient computation. We employed the AdamW optimizer with a learning rate of 1e-4 and a warmup ratio of 0.05. DeepSpeed ZeRO Stage 3~\cite{rasley2020deepspeed} was utilized for distributed training to optimize memory usage across GPUs. All training and evaluation tasks were implemented in PyTorch 2.4, and all experiments were performed on eight NVIDIA A100 GPUs with 80GB of memory each.

\noindent \textbf{Evaluation metric.}
We chose ROUGE-L as the evaluation metric because its recall calculation method, based on the longest common subsequence, can effectively tolerate redundant prefixes, format symbols, and other irrelevant content in the model output.

\subsection{Quantitative validation of multi-frame dependency}

We verify multi-frame dependency in ViTexQA from two perspectives: \textcolor{purple}{\textit{\textbf{Part I)}}} Explicit judgment on query-video pairs. We classify a query-video pair as multi-frame dependent or not, via annotators and prompted MLLMs. \textcolor{purple}{\textit{\textbf{Part II)}}} Implicit assessment via single/multi-frame performance gap. We evaluate four inputs: \textbf{Single-Rand} (a random video frame), \textbf{Single-AnsRand} (a random answer frame), \textbf{Single-Oracle} (the answer frame with the richest text), and \textbf{Multi-frame} (Full video). If models fail under single-frame input but succeed with multi-frame input, we regard query as multi-frame dependent.

\vspace{-16pt}
\begin{table*}[h]
\centering
\caption{Quantitative validation of multi-frame dependency in ViTexQA.}
\vspace{-1em}
\label{tab:dependency}
\scalebox{0.85}{
\begin{tabular}{lcccc}
\toprule
\rowcolor{blue!06}
\multicolumn{5}{c}{\textcolor{purple}{\textit{\textbf{Part~\uppercase\expandafter{\romannumeral1}: Explicit Judgment}}}}\\
% \multicolumn{4}{c}{\textbf{Part~\uppercase\expandafter {\romannumeral2}: Explicit Judgment}} \\
% \midrule
% \rowcolor{blue!03}
\textbf{Method} & \multicolumn{4}{c}{\textbf{The ratio of multi-frame dependent query-video pairs}} \\
% \rowcolor{blue!03}
Human & \multicolumn{4}{c}{\hl{100.0}}\\
% \rowcolor{blue!03}
Gemini-3 Pro & \multicolumn{4}{c}{\hl{95.4}} \\
% \rowcolor{blue!03}
Claude-Sonnet-4.6  & \multicolumn{4}{c}{\hl{97.7}} \\
\midrule
\rowcolor{green!06}
\multicolumn{5}{c}{\textcolor{purple}{\textit{\textbf{Part~\uppercase\expandafter{\romannumeral2}: Implicit Assessment}}}}\\
% \multicolumn{4}{c}{\textbf{Part~\uppercase\expandafter {\romannumeral1}: Implicit Evaluation}} \\
% \midrule
% \rowcolor{green!03}
\textbf{Method} & \textbf{Single-Rand} & \textbf{Single-AnsRand} & \textbf{Single-Oracle} & \textbf{Multi-frame} \\
% \textbf{Method} & \textbf{SF-Rand} & \textbf{SF-AnsRand} & \textbf{SF-Oracle} & \textbf{MF-Full} \\
% \rowcolor{green!03}
Gemini-3 Pro & 0.0 & 25.2& 27.3 & \hl{71.6} \\
% \rowcolor{green!03}
Qwen3-VL-8B  & 0.0 & 25.6& 26.5 & \hl{66.7} \\
\hline 
\end{tabular}}
\end{table*}
\vspace{-15pt}

As shown in Table~\ref{tab:dependency}, both parts consistently confirm the multi-frame dependency in ViTexQA: 1) Human annotators and MLLMs identify over 95\% of questions as multi-frame dependent. 2) Multi-frame input yields a dramatically higher score, confirming that a single frame is insufficient to answer the question and highlighting the multi-frame nature of ViTexQA.

\subsection{Effectiveness of Our ViTexQA Training Data}

% To validate the quality and generalizability of the ViTexQA training data, we fine-tune three representative models (Qwen3-VL-8B, MiniCPM-V4.5-8B, and FrameThinker-8B) on the ViTexQA training data and evaluate them across four benchmarks: ViTexQA benchmark, MME-VideoOCR, Video-MME, and TextVQA. As shown in Table~\ref{tab:finetune}, fine-tuning on the ViTexQA training data can significantly improve the three models' performance on the ViTexQA benchmark (by 10.2\%, 7.1\%, and 10.0\%, respectively), demonstrating the effectiveness of our curated training data for video text understanding tasks.

% More importantly, models fine-tuned on ViTexQA also achieve 
% improvements on out-of-domain benchmarks, including MME-VideoOCR, Video-MME, and TextVQA. This cross-benchmark generalization suggests that ViTexQA captures fundamental capabilities related to multi-frame textual perception, rather than merely overfitting to dataset-specific patterns. These results confirm that ViTexQA serves not only as a challenging evaluation benchmark but also as a high-quality training resource that can broadly enhance a model's ability to integrate and reason over textual information distributed across video frames. Furthermore, the results from TextVQA show that it also plays a positive role in extracting text information from single images.

To validate the quality and generalizability of the ViTexQA training data, we fine-tune three representative models on the ViTexQA training data and evaluate them across four benchmarks. As shown in Table~\ref{tab:finetune}, fine-tuning on the ViTexQA training data can significantly improve the three models' performance on the ViTexQA benchmark (by 10.2\%, 7.1\%, and 10.0\%, respectively), demonstrating the effectiveness of our curated training data for video text understanding tasks.

\vspace{-10pt}
\begin{table*}[h]
\centering
\caption{Performance comparison before and after SFT on ViTexQA training data across multiple benchmarks. \textbf{VTD.}: ViTexQA training data.}
\vspace{-1em}
\label{tab:finetune}
\scalebox{0.85}{ 
% \begin{tabular*}{\textwidth}{@{\extracolsep{\fill}}llllll}
\begin{tabular}{lcllll}  % 改用tabular
\hline
\textbf{Method} & \textbf{VTD.} & \textbf{ViTexQA} & \textbf{MME-VideoOCR}  & \textbf{Video-MME}  & \textbf{TextVQA}  \\
\hline
\multirow{2}{*}{Qwen3-VL}     
    & \usym{2717}  & 66.7 & 68.5 & 71.4 & 84.5 \\
    & \cellcolor[HTML]{F2F9FD}\usym{2713}  
    & \cellcolor[HTML]{F2F9FD}\textbf{76.9}{\tiny\textcolor[HTML]{487DB8}{\textbf{($+$10.2)}}} 
    & \cellcolor[HTML]{F2F9FD}\textbf{74.3}{\tiny\textcolor[HTML]{487DB8}{\textbf{($+$5.8)}}} 
    & \cellcolor[HTML]{F2F9FD}\textbf{72.6}{\tiny\textcolor[HTML]{487DB8}{\textbf{($+$1.2)}}} 
    & \cellcolor[HTML]{F2F9FD}\textbf{84.6}{\tiny\textcolor[HTML]{487DB8}{\textbf{($+$0.1)}}} \\
\hline
\multirow{2}{*}{MiniCPM-V4.5} 
    & \usym{2717}  & 70.1 & 77.5 & 68.8 & 82.4 \\
    & \cellcolor[HTML]{F2F9FD}\usym{2713}  
    & \cellcolor[HTML]{F2F9FD}\textbf{77.2}{\tiny\textcolor[HTML]{487DB8}{\textbf{($+$7.1)}}}
    & \cellcolor[HTML]{F2F9FD}\textbf{82.3}{\tiny\textcolor[HTML]{487DB8}{\textbf{($+$4.8)}}}
    & \cellcolor[HTML]{F2F9FD}\textbf{70.6}{\tiny\textcolor[HTML]{487DB8}{\textbf{($+$1.6)}}}
    & \cellcolor[HTML]{F2F9FD}\textbf{82.8}{\tiny\textcolor[HTML]{487DB8}{\textbf{($+$0.4)}}} \\
\hline
\multirow{2}{*}{FrameThinker} 
    & \usym{2717}  & 73.5 & 78.3 & 74.1 & 84.2 \\
    & \cellcolor[HTML]{F2F9FD}\usym{2713}   
    & \cellcolor[HTML]{F2F9FD}\textbf{83.5}{\tiny\textcolor[HTML]{487DB8}{\textbf{($+$10.0)}}}
    & \cellcolor[HTML]{F2F9FD}\textbf{88.1}{\tiny\textcolor[HTML]{487DB8}{\textbf{($+$9.8)}}}
    & \cellcolor[HTML]{F2F9FD}\textbf{75.3}{\tiny\textcolor[HTML]{487DB8}{\textbf{($+$1.2)}}}
    & \cellcolor[HTML]{F2F9FD}\textbf{85.3}{\tiny\textcolor[HTML]{487DB8}{\textbf{($+$1.1)}}} \\
\hline 
\end{tabular}}
\vspace{-2mm}
\scalebox{0.85}{
\begin{minipage}{\textwidth}
\small
$^*$ All models use CoT during inference, regardless of whether SFT is performed.
\end{minipage}
}
\vspace{-1em}
\end{table*}

More importantly, models fine-tuned on ViTexQA also achieve 
improvements on out-of-domain benchmarks. These results confirm that ViTexQA serves not only as a challenging evaluation benchmark but also as a high-quality training resource that can broadly enhance a model's ability to integrate and reason over textual information distributed across video frames. Furthermore, the results from TextVQA show that it also plays a positive role in extracting text information from single images.

\subsection{Effectiveness of Our FrameThinker on ViTexQA.}

\subsubsection{Zero-Shot Comparison with Existing MLLMs.} To evaluate the effectiveness of FrameThinker, we first conduct a fair comparison with existing MLLMs under a zero-shot setting, where no ViTexQA training data is utilized by any of the compared methods. The corresponding results are presented in Table~\ref{tab:result}. FrameThinker, trained on the custom multi-frame perception CoT data from M4-ViteVQA, achieves the best overall performance with an average score of 73.5\%, outperforming all competing baselines by a significant margin.

% \vspace{-10pt}
\begin{table*}[t]
\centering
\caption{Performance of existing MLLMs in zero-shot setting on the ViTexQA benchmark. \textbf{Avg.}:the average performance of the 30 scenarios; \textbf{Dri.}:Driving; \textbf{Adv.}:Advertising; \textbf{Edu.}:Education; \textbf{RT.}:RollingText. The highest score are highlighted, and the second highest is underlined. }
\label{tab:result}
\vspace{-1em}
\scalebox{0.85}{ 
% \begin{tabular*}{\textwidth}{@{\extracolsep{\fill}}lcccccccc@{}}
\begin{tabular}{lcccccccc}  % 改用tabular
% \begin{tabular*}{\textwidth}{@{}@{\extracolsep{\fill}}lcccccccc@{}@{}}
\hline
\textbf{Method} & \textbf{Size} & \textbf{Max\_Frames} & \textbf{Resolution} & \textbf{Avg.} & \textbf{Dri.} & \textbf{Adv.} & \textbf{Edu.} & \textbf{RT.} \\
\hline
% \rowcolor{green!06} 
\multicolumn{9}{c}{Closed-source MLLMs} \\
\hline
% \rowcolor{green!06} 
GPT-4o~\cite{openai2024}       &- &128&512*512& 51.6&55.2&41.4&51.9&46.6 \\
% \rowcolor{green!06} 
Gemini-3 Pro~\cite{gemini25}  & -  &128&512*512& 71.6&\textbf{70.9}&70.3&58.9&59.2  \\
% \rowcolor{green!06} 
Seed-1.6~\cite{guo2025seed1} & -  &128&512*512& 61.1&63.4&60.2&54.9&56.8 \\
\hline
% \rowcolor{blue!06} 
\multicolumn{9}{c}{Open-source MLLMs} \\
\hline
% \rowcolor{blue!06} 
Qwen3-VL~\cite{qwen3}       & 2B  & 768      & 448*448 & 57.2    & 56.9   & 56.9&    69.1  &52.4    \\
% \rowcolor{blue!06}
LongVU~\cite{shen2024longuv}           & 3B  &768      & 448*448&  43.1&45.3&47.5&46.5&36.4  \\
% \rowcolor{blue!06}
Qwen2.5-VL~\cite{bai2025qwen2}      & 3B  & 768      & 448*448& 55.4& 58.1& 53.3& 63.2& 49.4         \\
% \rowcolor{blue!06}
Qwen3-VL~\cite{qwen3}       & 4B  &768      & 448*448&62.3&62.1&59.0&71.7&57.7      \\
% \rowcolor{blue!06}
MiniCPM-V2.6~\cite{yao2024minicp}   & 7B  &64      & 448*448& 50.7&55.6&51.7&57.0&38.2   \\
% \rowcolor{blue!06}
Qwen2.5-VL~\cite{bai2025qwen2}     & 7B  &768      & 448*448&63.5&64.4&63.9&73.3&58.5   \\
% \rowcolor{blue!06}
VideoLLaMA3~\cite{zhang2025videollama3}    & 7B &64   & 336*336& 52.4&54.5&62.1&58.5&49.6 \\
% \rowcolor{blue!06}
LLaVA-Video~\cite{zhang2025llavavi}    & 7B   &16      & 336*336& 36.0&58.3&46.3&41.0&37.6   \\
% \rowcolor{blue!06}
Keye-VL~\cite{yang2025kwai}         & 8B    &768      & 336*336&  39.9&54.9&47.1&49.1&35.3                \\
% \rowcolor{blue!06}
MiniCPM-V4.5~\cite{yao2024minicp}   & 8B   &768      & 448*448&  \underline{70.1}&48.6&77.4&51.1& \underline{60.1}                       \\
% \rowcolor{blue!06}
Qwen3-VL~\cite{qwen3}       & 8B  &768      & 448*448& 66.7&63.7&61.7&\underline{73.6}& 53.4  \\
% \rowcolor{blue!06}
InternVL3~\cite{zhu2025internv}   & 8B  &64      & 448*448& 64.2&52.1&\underline{85.5}&60.0&52.3 \\
\hline
% \arrayrulecolor{blue!06}
 \rowcolor[HTML]{E5F4FB}
\textbf{FrameThinker}   & 8B  &768      & 448*448& \textbf{73.5}&\underline{67.4}& \textbf{87.2}&\textbf{75.9}&\textbf{62.3}\\
\hline
\end{tabular}}
\vspace{-1em}
\end{table*}

Table~\ref{tab:result} shows that the open-source models have a clear technical advantage, significantly outperforming closed-source models. This superior performance is mainly due to the resource limitation of a maximum input of 128 frames, while the open-source model can utilize more frames, which reflects the ViTexQA dataset's emphasis on the core capabilities of model integration and inference of multi-frame content.

At the same time, all models perform worst on Rolling Text, revealing a fundamental weakness in current methods regarding the integration of dynamically changing textual information across frames. This represents the core evaluation challenge that ViTexQA is designed to assess.

% Furthermore, FrameThinker achieved an average score of 73.5, while Keye-VL ~\cite{yang2025kwai}, using the same 8B parameter set, scored only 39.9. This demonstrates that model performance is not linearly correlated with parameter size, suggesting that task-specific model design and training strategies are more critical than simply increasing the number of model parameters.

\noindent \textbf{Effectiveness of FrameThinker Trained on ViTexQA Data.} To further evaluate the effectiveness of FrameThinker when trained on ViTexQA data, we present a comprehensive comparison in Table~\ref{tab:Ablation2}, contrasting the results before and after incorporating ViTexQA training data. As shown in the table, \textbf{No.(2)}, \textbf{No.(3)}, and \textbf{No.(4)} achieve improvements of 2.8\%, 7.9\%, and 10.0\% over \textbf{No.(1)}, respectively.

% To further illustrate the challenges posed by ViTexQA, Fig.~\ref{fig:qualitative} presents qualitative results of representative models (Gemini-3 Pro, Seed-1.6, MiniCPM-V4.5, and FrameThinker) on both real-world and synthetic videos from ViTexQA. As shown in the figure, all models struggle to provide accurate answers, even the best-performing models in quantitative evaluation. This difficulty stems from the fundamental requirement of integrating textual information scattered across multiple frames.

% \begin{figure*}[h]
%     \centering
%     \includegraphics[width=0.95\linewidth]{sec/img/example.png}
%     \caption{Qualitative experimental results of representative models on ViTexQA.}
%     \label{fig:qualitative}
% \end{figure*}

% F2F9FD E5F4FB

\subsection{Ablation studies}

% \subsubsection{Each Subset results of ViTexQA}

\subsubsection{The necessity of multi-frame perception.}

% To investigate the impact of video input resolution and frame count on the ViTexQA dataset, we conducted two sets of experiments. Fig.~\ref{fig:Ablation1} (a) illustrates the resolution experiment, in which the maximum input frame count was set to 768. As the resolution increased, model performance also improved; however, after reaching $448^2$, further increases in resolution yielded marginal gains. Fig.~\ref{fig:Ablation1} (b) presents the experiment on maximum frame count, where the resolution was fixed at $448^2$. In this experiment, model performance continued to improve as the input frame count increased.

\begin{figure*}[h]
    \centering
    \includegraphics[width=0.95\linewidth]{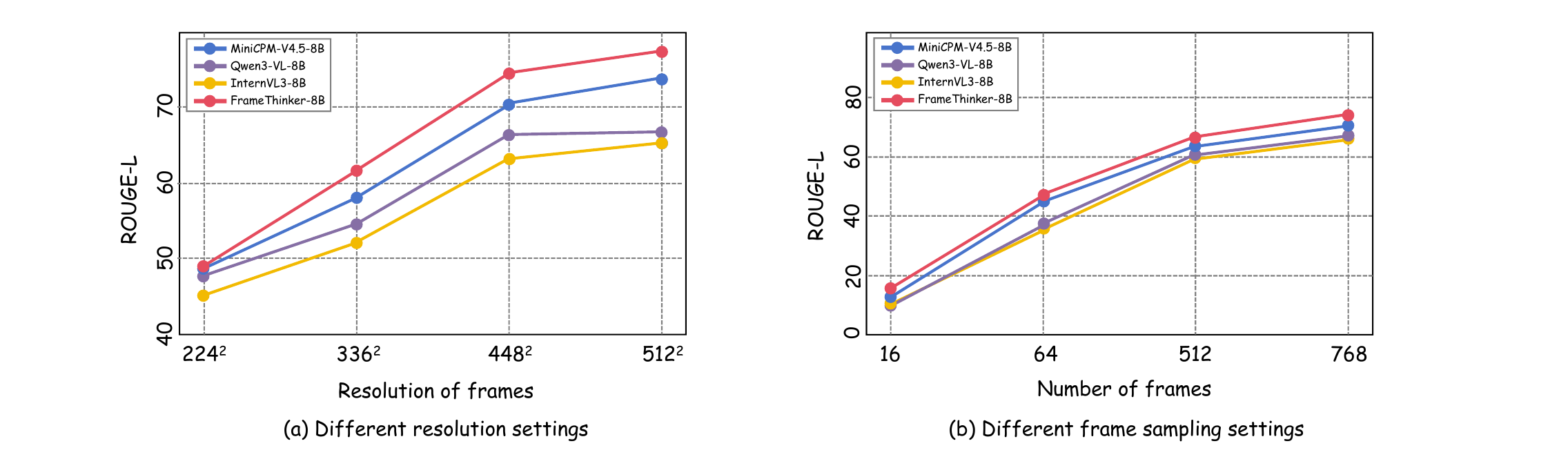}
    \vspace{-1em}
    \caption{The model's performance on ViTexQA is compared under different resolutions and frame sampling settings.}
    \label{fig:Ablation1}
\end{figure*}

To analyze how video resolution and frame number affect performance on ViTexQA, we perform two ablation studies. Fig.~\ref{fig:Ablation1}(a) fixes the maximum frame count to 768 for resolution evaluation: accuracy rises with resolution, yet gains become negligible beyond $448^2$. Fig.~\ref{fig:Ablation1}(b) uses a fixed $448^2$ resolution to test frame counts, showing consistent performance improvements with more input frames.

\vspace{-15pt}
\vspace{-10pt}

% \begin{table}[h]
% \centering
% % \footnotesize
% \caption{Ablation Study on the gradient sampling.}
% \label{tab:Ablation1}
% \resizebox{0.95\linewidth}{!}{
% \begin{tabular*}{\linewidth}{@{\extracolsep{\fill}}lcccccc}
% \toprule
% \textbf{Frame Sampling Ratio} & \textbf{10\%} & \textbf{30\%} & \textbf{50\%} & \textbf{70\%} & \textbf{90\%}& \textbf{100\%} \\
% \midrule
% ViTexQA\#Test & 16.4 & 24.6 & 39.7 & 48.2 & 59.1 & \textbf{66.7} \\
% \bottomrule
% \end{tabular*}}
% \end{table}

\noindent \textbf{Ablation studies on FrameThinker.} Table~\ref{tab:Ablation2} presents a series of ablation experiments, demonstrating that FrameThinker's CoT-guided SFT and Temporally-grounded RL can improve the model's perception ability for multi-frame video text question answering tasks. As shown in the table, \textbf{No.(3)}, and \textbf{No.(4)} achieve improvements of 5.1\% and 7.2\%, over \textbf{No.(2)}, respectively.

\vspace{-15pt}
\begin{table*}[h]
\centering
% \caption{Performance of FrameThinker before and after using ViTexQA training data and ablation study on the FrameThinker. \textbf{No.(1)} represents the baseline without SFT on ViTexQA training data. \textbf{No.(2)} denotes the result of SFT solely on the ViTexQA training data without additional perception supervision. \textbf{No.(3)} corresponds to the result of SFT with our constructed multi-frame perception CoT data derived from the ViTexQA training set. \textbf{No.(4)} represents the result of our FrameThinker.}
\caption{Performance of FrameThinker before and after using ViTexQA training data and ablation study on the FrameThinker.}
\vspace{-1.em}
\label{tab:Ablation2}
\scalebox{0.85}{ 
\begin{tabular}{c|cccllll}  % 改用tabular
\hline
\textbf{No.} & \textbf{VTD.} & \textbf{CoT-SFT}& \textbf{RL} & \textbf{ViTexQA} & \textbf{MME-VideoOCR}  & \textbf{Video-MME}  & \textbf{TextVQA}  \\
\hline
\textbf{(1)} & \usym{2717} & \usym{2717} & \usym{2717}  & 73.5 & 78.3 & 74.1 & 84.2 \\

\textbf{(2)}& \cellcolor[HTML]{E9F8FE}\usym{2713}
& \cellcolor[HTML]{E9F8FE}\usym{2717} 
& \cellcolor[HTML]{E9F8FE}\usym{2717}  
& \cellcolor[HTML]{E9F8FE}76.3{\tiny\textcolor[HTML]{487DB8}{\textbf{($+$2.8)}}}
& \cellcolor[HTML]{E9F8FE}80.5{\tiny\textcolor[HTML]{487DB8}{\textbf{($+$2.2)}}}
& \cellcolor[HTML]{E9F8FE}74.4{\tiny\textcolor[HTML]{487DB8}{\textbf{($+$0.3)}}}
& \cellcolor[HTML]{E9F8FE}84.2{\tiny\textcolor[HTML]{487DB8}{\textbf{($+$0.0)}}} \\
\textbf{(3)}& \cellcolor[HTML]{D9F2FD}\usym{2713}  
& \cellcolor[HTML]{D9F2FD}\usym{2713} 
& \cellcolor[HTML]{D9F2FD}\usym{2717}  
& \cellcolor[HTML]{D9F2FD}81.4{\tiny\textcolor[HTML]{487DB8}{\textbf{($+$7.9)}}}
& \cellcolor[HTML]{D9F2FD}85.3{\tiny\textcolor[HTML]{487DB8}{\textbf{($+$7.0)}}}
& \cellcolor[HTML]{D9F2FD}75.2{\tiny\textcolor[HTML]{487DB8}{\textbf{($+$1.1)}}} 
& \cellcolor[HTML]{D9F2FD}84.9{\tiny\textcolor[HTML]{487DB8}{\textbf{($+$0.7)}}} \\
\textbf{(4)}& \cellcolor[HTML]{C6EBFA}\usym{2713}  
& \cellcolor[HTML]{C6EBFA}\usym{2713} 
& \cellcolor[HTML]{C6EBFA}\usym{2713}
    & \cellcolor[HTML]{C6EBFA}\textbf{83.5}{\tiny\textcolor[HTML]{487DB8}{\textbf{($+$10.0)}}}
    & \cellcolor[HTML]{C6EBFA}\textbf{88.1}{\tiny\textcolor[HTML]{487DB8}{\textbf{($+$9.8)}}}
    & \cellcolor[HTML]{C6EBFA}\textbf{75.5}{\tiny\textcolor[HTML]{487DB8}{\textbf{($+$1.4)}}}
    & \cellcolor[HTML]{C6EBFA}\textbf{85.3}{\tiny\textcolor[HTML]{487DB8}{\textbf{($+$1.1)}}} \\
\hline
\end{tabular}}
% \vspace{-2em}
\end{table*}
\vspace{-20pt}

\section{Conclusion}

% In this work, we introduce ViTexQA, a large-scale multi-frame temporal reasoning dataset specifically designed for video text question answering, addressing the gap in temporal text reasoning benchmarks. Through a comprehensive evaluation of mainstream MLLMs on ViTexQA, we found that they have significant performance differences in cross-frame text fusion, and their overall performance is poor. To address these challenges, we propose FrameThinker, a multi-modal training and reasoning strategy. By leveraging explicit text extraction from key frames and enhanced reasoning mechanisms, FrameThinker substantially improves performance. Our work demonstrates that while existing MLLMs struggle with temporal video text reasoning, task-specific architectural innovations combined with domain-adapted fine-tuning can effectively bridge this gap, opening new directions for advancing video understanding in the era of large language models.

In this paper, we introduce ViTexQA, a large-scale multi-frame temporal perception dataset for video text question answering, addressing the gap in temporal text perception datasets and benchmarks. Evaluations of mainstream MLLMs on ViTexQA reveal weak cross-frame text fusion and unsatisfactory overall performance. To address these challenges, we propose FrameThinker to boost multi-frame temporal perception via explicit key-frame text extraction and optimized perception modules, achieving clear performance gains. We believe that the dataset, multi-frame perception CoT construction strategy, and training framework introduced in this work will serve as valuable resources for the research community, stimulating further progress toward MLLMs that can genuinely comprehend the rich, temporally distributed textual information present in real-world videos.

\bibliographystyle{splncs04}
\bibliography{main}

\clearpage
\setcounter{page}{1}
% \maketitlesupplementary

% \begin{center}
%     {\large Supplementary Material}
% \end{center}

\appendix

\section{Overview of Appendix}

\begin{itemize}
\item Collecting Details of ViTexQA.
\item Details of Annotation.
\item Representative Examples from ViTexQA.
\item Experiment Details.
\item Broader Impact.
\end{itemize}

\section{Collecting Details of ViTexQA}
\begin{figure*}[h]
    \centering
    \includegraphics[width=0.95\linewidth]{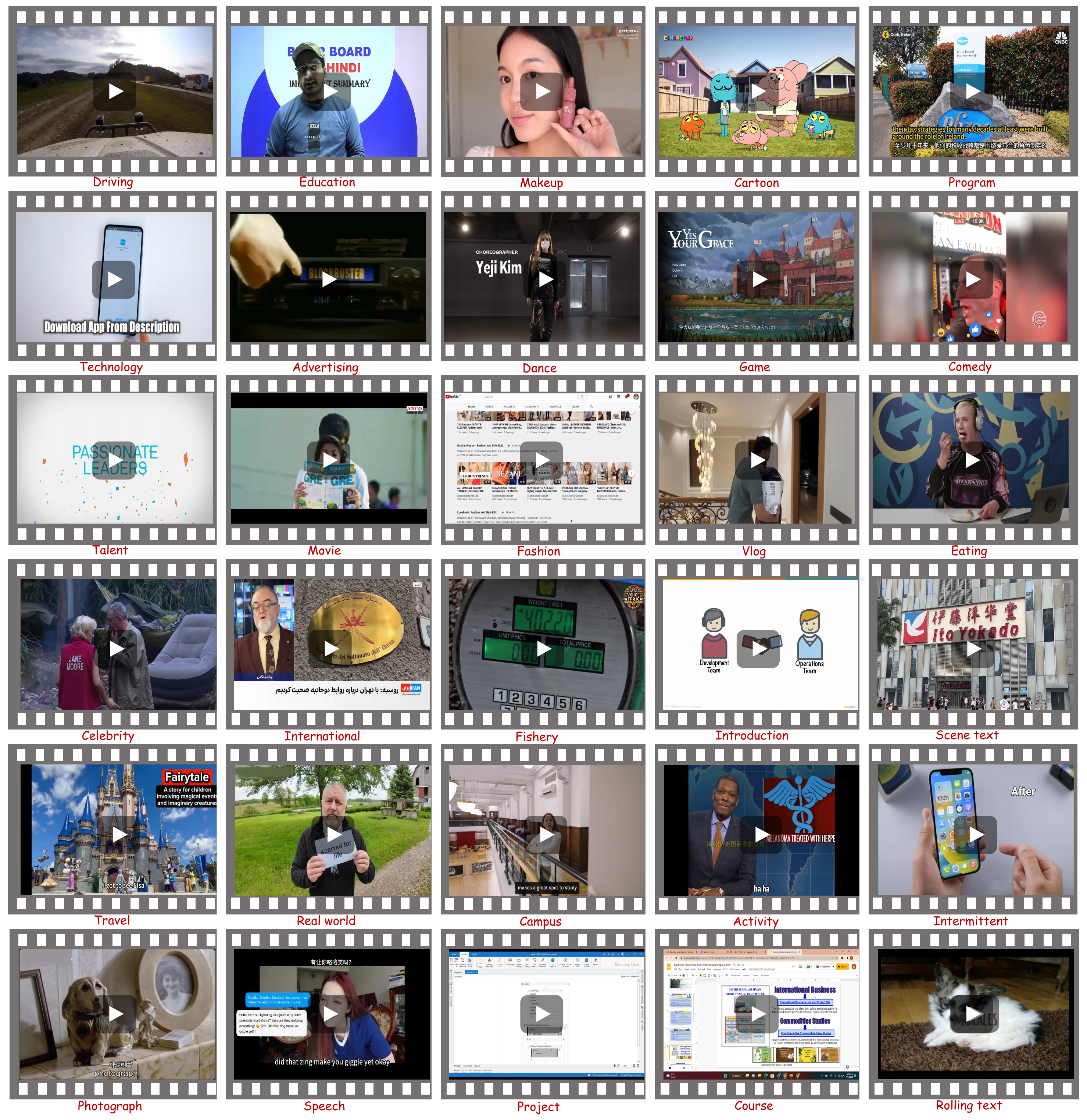}
    \caption{Video demonstrations of various scenarios in the ViTexQA dataset.}
    \label{fig:youtube}
\end{figure*}

This section provides a comprehensive overview of the acquisition, filtering, synthesis, and analysis processes employed to construct the video content in ViTexQA, ensuring both semantic richness and temporal complexity across diverse real-world scenarios. To establish a robust foundation for temporal textual inference, we collected 11,924 videos from YouTube, focusing on categories that naturally contain rich, temporally distributed textual information across multiple frames. We also collected 1,000 challenging videos from existing datasets with the original annotations removed. In addition, to enhance the coverage of temporal textual patterns, we synthesized 100 Rolling-text videos using synthetic methods, as shown in Fig.~\ref{fig:youtube}.

\subsection{Real Video Acquisition}

\subsubsection{YouTube Video Retrieval and Filtering Criteria}
Candidate videos were retrieved using category-specific keyword queries designed to target text-rich content, such as ``Education\_Text'', ``Game\_OCR'', ``Travel\_Content'', ``News\_ Subtitle'', ``Sports\_Score'' and ``Fashion\_Product''. To ensure dataset quality and suitability for multi-frame text perception tasks, we implemented the following stringent filtering rules:

\begin{itemize}
    \item \textbf{Duration Range.} Videos were constrained to a duration between 1 and 30 minutes. This range balances temporal complexity with computational efficiency, ensuring sufficient frame sequences for cross-frame perception while maintaining manageable processing requirements.
    \item \textbf{Resolution Requirements.} To guarantee text clarity and readability across frames, we established a minimum resolution threshold of 720p. Videos with resolutions below this threshold were systematically discarded to prevent text degradation that could compromise annotation quality. Higher-resolution videos were downloaded directly at 720p, ensuring uniform visual quality across the entire dataset. This standardization facilitates efficient storage, processing, and fair model evaluation.
    \item \textbf{Scene-Text Richness.} We employed OCR engine~\cite{du2020pp}, a state-of-the-art optical character recognition system, to assess the textual density of each candidate video. Videos containing fewer than 30 detected text words were excluded from the dataset. This threshold ensures that retained videos contain sufficient textual information distributed across frames to support meaningful temporal perception tasks. The OCR-based filtering process involved sampling frames at regular intervals (every 5 seconds) and aggregating detected text instances to estimate overall textual richness.
\end{itemize}

\subsubsection{Existing Dataset Video Selection}
To further enrich the diversity and challenge level of ViTexQA, we curated an additional collection of 1,000 videos from several established benchmark datasets, including NewsVideoQA~\cite{jahagirdar2023watching}, ICDAR2013~\cite{karatzas2013icdar}, and ICDAR2023~\cite{wu2023icdar}. The selection process specifically targeted videos whose original annotations involved complex, multi-instance, or densely distributed textual content, as such videos inherently demand sophisticated temporal and spatial text perception capabilities. All original annotations associated with the selected videos were deliberately discarded to prevent annotation leakage and to ensure that our newly constructed question-answer (QA) pairs reflect the specific demands of temporal textual inference rather than inheriting the task formulations of the source datasets.

\subsubsection{Metadata Statistics}
Beyond visual content, we systematically collected comprehensive metadata for each video, including duration, fps, total frame count, and aspect ratio. This metadata enables detailed analysis of coverage diversity across temporal and visual characteristics. Statistical analysis reveals that our dataset exhibits substantial variation in temporal properties: video durations range from 10 seconds to 30 minutes (mean: 6.32 minutes), frame rates vary between 24 and 60 fps (with 30 fps being most common), and total frame counts span from 360 to 54,000 frames. This heterogeneity ensures that models trained and evaluated on ViTexQA must handle diverse temporal dynamics and frame sampling strategies, better reflecting real-world video understanding scenarios.

\subsection{Rolling-Text Video Synthesis}
\begin{figure*}[t]
    \centering
    \includegraphics[width=0.95\linewidth]{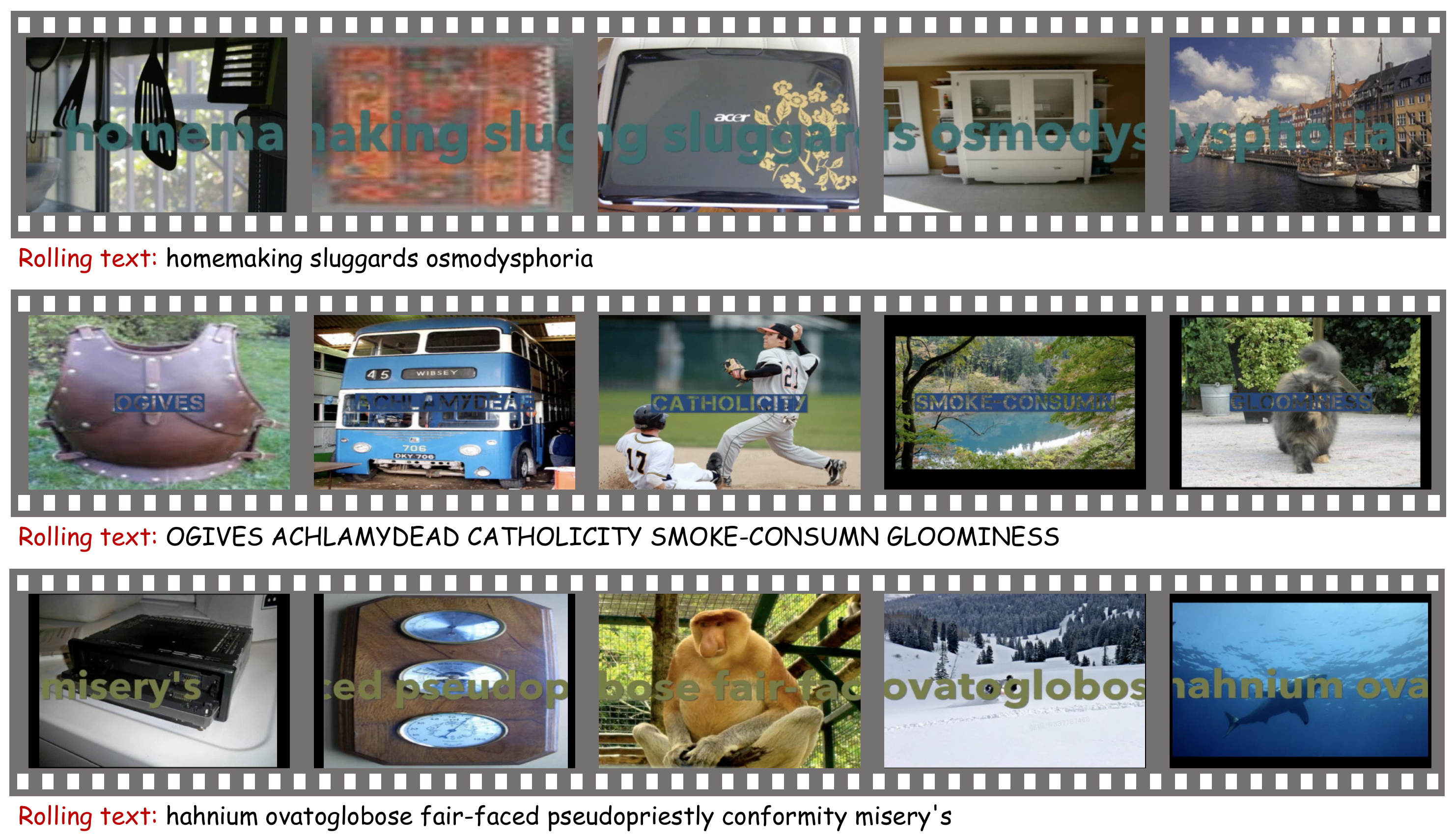}
    \caption{Video demonstration of synthesized ``Rolling\_text'' scenarios from the ViTexQA dataset.}
    \label{fig:sys}
\end{figure*}

To further enrich the dataset's coverage of multi-frame temporal text patterns and ensure controlled evaluation scenarios, we developed an in-house Rolling-Text video synthesis pipeline that generates 100 synthetic videos, as shown in Fig.~\ref{fig:sys}. These videos are specifically designed such that textual information is not isolated within single frames but rather unfolds continuously, segmentally, or as supplementary explanations across multiple frames. Consequently, information from any single frame is insufficient to comprehend the complete semantic content, necessitating genuine cross-frame integration.

\subsubsection{Synthesis Pipeline and Parameter Diversity}
The synthesis process begins with curating a collection of text-free background images spanning diverse visual contexts. These backgrounds are then assembled into video sequences using FFmpeg, with the following randomized textual and visual elements integrated to maximize variability and realism:

\begin{itemize}
    \item \textbf{Typography and Linguistic Diversity.} We incorporated over 20 distinct font families ranging from serif to sans-serif, monospace, and decorative styles. Textual content was sampled from the COCA (Corpus of Contemporary American English) word corpus to ensure linguistic authenticity and semantic diversity. Additional parameters include font color, opacity levels, Gaussian noise application, shadow size, shadow color and transparency, background overlay colors, and background transparency. These variations ensure that synthesized text exhibits visual characteristics comparable to naturally occurring scene text.
    \item \textbf{Animation and Motion Patterns.} Text appearance follows diverse motion patterns including left-to-right scrolling, right-to-left scrolling, top-to-bottom vertical motion, bottom-to-top ascension, typewriter-style character-by-character revelation, fade-in/fade-out transitions, and zoom effects. Frame transition animations include dissolve, wipe, and slide effects to simulate realistic video editing techniques. The temporal distribution of text ensures that complete semantic units emerge only through observing consecutive frames, directly enforcing the temporal perception requirement central to ViTexQA.
\end{itemize}

\subsubsection{Metadata Statistics}
Due to the randomized nature of synthesis parameters, each generated video exhibits unique temporal characteristics. Video durations range from 30 to 600 seconds, frame rates vary between 24 and 30 fps, and total frame counts span from 720 to 18,000 frames. Critically, all synthesis parameters—including text content, timing, motion trajectories, and visual attributes—are logged during generation, providing ground-truth annotations that facilitate precise evaluation of temporal text extraction and perception capabilities. This synthetic subset serves as a controlled testbed for isolating specific challenges in multi-frame text integration.

\subsection{Word Cloud Distribution Analysis}
\begin{figure*}[h]
    \centering
    \includegraphics[width=0.95\linewidth]{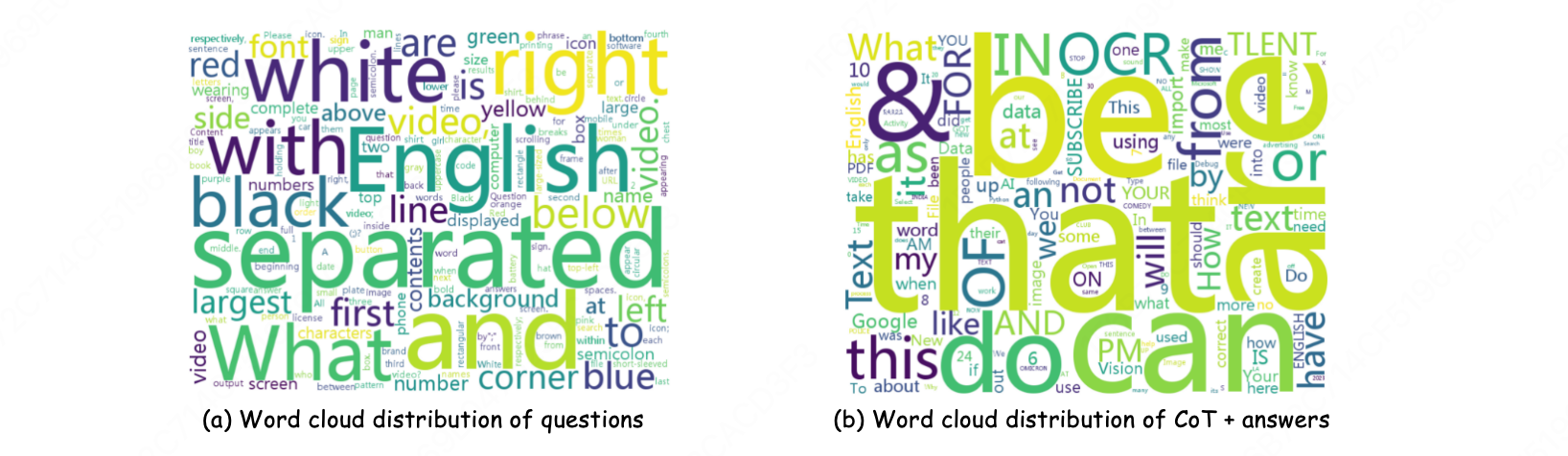}
    \caption{Word cloud distribution in ViTexQA dataset.}
    \label{fig:ciyun}
\end{figure*}

To visualize and quantitatively assess the semantic richness and diversity of textual interactions within ViTexQA, we constructed a word cloud representation based on the 200 most frequently occurring terms across all questions and answers in the dataset. As illustrated in Fig.~\ref{fig:ciyun}, high-frequency terms such as ``English,'' ``video,'' ``text,'' and ``content'' reflect the strong alignment between textual elements and semantic perception requirements inherent to the dataset. The prominent presence of spatial keywords including ``left,'' ``right,'' underscores the importance of spatial awareness in video text understanding, while functional descriptors including ``white,'' ``black,'' highlight the multi-granular nature of perception demands from low-level visual attribute recognition to high-level semantic comprehension. Furthermore, the co-occurrence of contextual separators and logical connectors including ``separated,'' ``and,'' emphasizes the temporal and relational perception aspects central to ViTexQA. The word cloud analysis reveals that questions span a wide spectrum of cognitive complexities, requiring models to integrate information across frames, recognize temporal sequences, perform spatial perception, and synthesize multi-modal cues.

\section{Details of Annotation}
This section elaborates on the Quality-Controlled annotationmethodology employed to construct ViTexQA.

\subsection{Annotation of Real Video Instances}
In stark contrast to prevalent practices that rely on initial machine-generated annotations followed by human verification, our annotation process was conducted entirely by trained human annotators, ensuring authenticity and minimizing biases inherent to model-generated content. The annotation team comprised eight dedicated annotators and four independent evaluators who collectively invested three months in producing high-quality QA pairs.

The overall annotation workflow was structured into three iterative rounds, each serving distinct quality assurance objectives, as detailed in Table~\ref{tab:duration}:

\begin{table*}[h]
\centering
\caption{Annotation workflow of ViTexQA real videos.}
\label{tab:duration}
\resizebox{0.95\linewidth}{!}{
\begin{tabular}{lccccc}
\toprule
\multirow{2}{*}{\textbf{Rounds}} & \multirow{2}{*}{\textbf{Task}} & \multirow{2}{*}{\textbf{Task duration}} & \textbf{Number of} & \textbf{Number of} & \textbf{Number of} \\
& & & \textbf{Videos} & \textbf{Q\&A pairs} & \textbf{members} \\
\midrule
Round 1 & Initial Generation & 2025-05-08 $\to$ 2025-07-17 & 12924 $\to$ 8925 & 0 $\to$ 20867 & 8 \\
Round 2 & Quality Scoring & 2025-07-18 $\to$ 2025-07-28 & 8925 $\to$ 5280 & 20867 $\to$ 6902 & 4 \\
Round 3 & Revision and Return Scoring & 2025-07-29 $\to$ 2025-08-15 & 5280 $\to$ 5047 & 6902 $\to$ 6764 & 4 \\
\bottomrule
\end{tabular}}
\end{table*}

\begin{figure*}[t]
    \centering
    \includegraphics[width=0.99\linewidth]{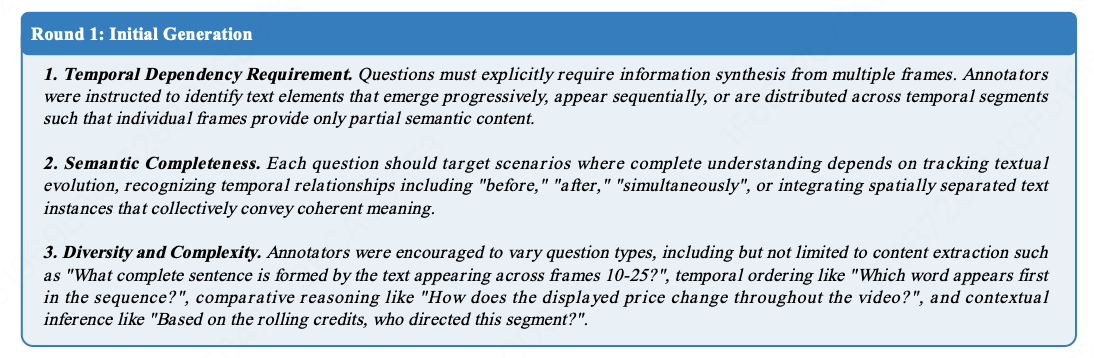}
    \caption{Comprehensive limitations of Initial Generation.}
    \label{fig:guideline1}
\end{figure*}

\noindent \textbf{Round 1: Initial Generation.} Annotators were tasked with carefully examining each video to formulate semantically meaningful QA pairs tightly coupled with OCR-detectable text appearing across video frames. Annotators received the comprehensive limitations in Fig.~\ref{fig:guideline1}.

\begin{figure*}[t]
    \centering
    \includegraphics[width=0.99\linewidth]{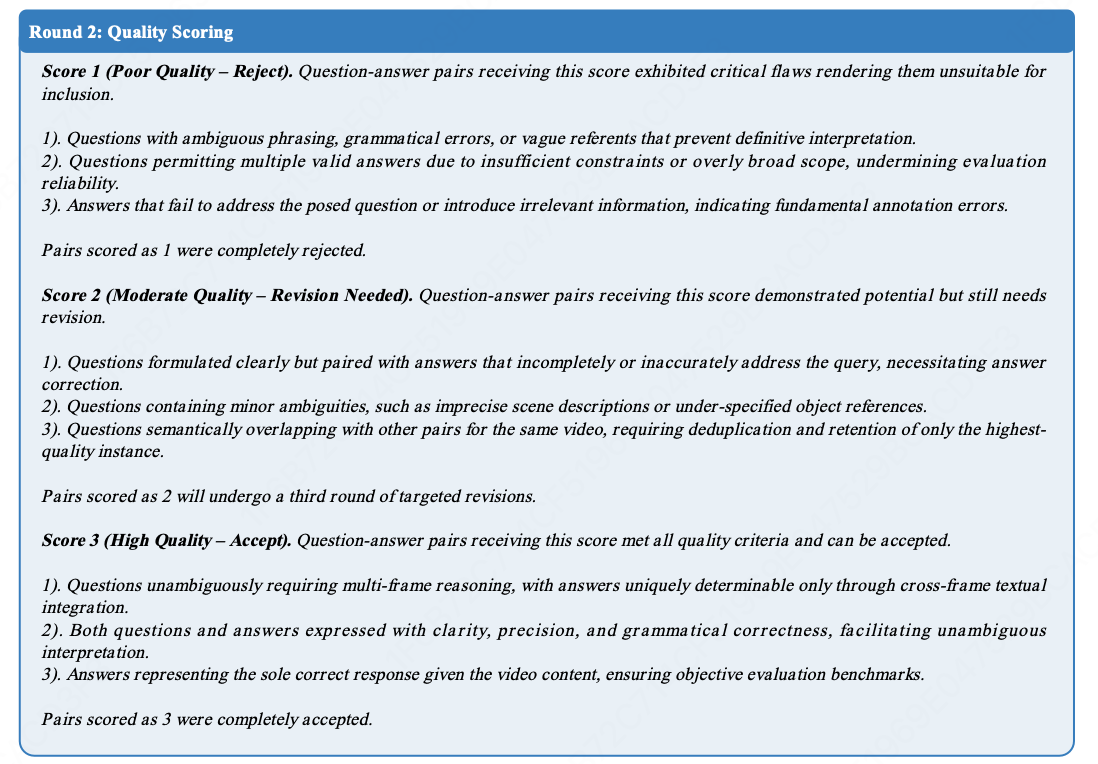}
    \caption{Three-level scoring of Quality Scoring.}
    \label{fig:guideline2}
\end{figure*}

\begin{figure*}[h]
    \centering
    \includegraphics[width=0.99\linewidth]{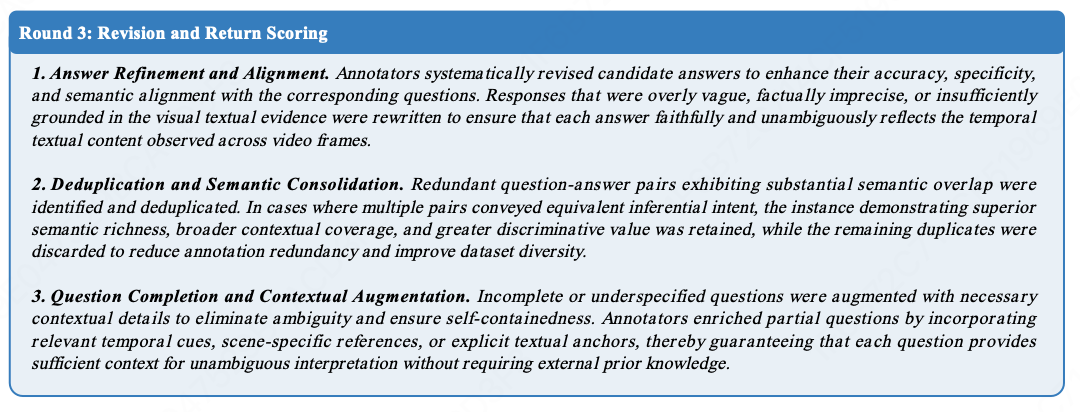}
    \caption{Modification limitations of Revision and Return Scoring.}
    \label{fig:guideline3}
\end{figure*}

This initial annotation phase yielded a raw collection of QA candidates that subsequently underwent rigorous evaluation to ensure quality and temporal perception alignment.

\noindent \textbf{Round 2: Quality Scoring.} To maintain stringent quality standards, four independent evaluators systematically reviewed all candidate QA pairs generated in Round 1. Each evaluator assesses the QA pairs according to the three-level scoring criteria in Fig.~\ref{fig:guideline2}, which are designed to identify and categorize QA pairs.

In this step, QA pairs scoring 1 point will be rejected, those scoring 2 points will proceed to the next round for revision, and those scoring 3 points will be accepted directly.

\noindent \textbf{Round 3: Revision and Return Scoring.} Based on the results of the second round of evaluation, the annotators made targeted modifications according to the rules of :

The limitations shown in Fig.~\ref{fig:guideline3} are revised for the QA pairs that received a score 2 in the second round. The revised QA pairs will then be returned to the second round for scoring.

This iterative three-round process, combining independent annotation with multi-evaluator review and targeted revision, ensures that ViTexQA comprises exclusively high-quality QA pairs that authentically enforce temporal perception requirements.

\subsection{Annotation of Synthetic Video Instances}
The annotation process for the 100 synthesized rolling-text videos leveraged the controlled nature of the synthesis pipeline to streamline annotation while maintaining consistency. Since all synthesis parameters including complete textual content, timing, motion trajectories, and visual attributes—were logged during video generation, ground-truth annotations were readily available.

\begin{itemize}
    \item \textbf{Answer Annotation.} The complete rolling or sequentially displayed text content synthesized for each video directly served as the answer, ensuring perfect alignment between ground truth and visual content.
    \item \textbf{Question Standardization.} To maintain uniformity and facilitate systematic evaluation, all questions for synthetic videos were standardized to: \texttt{``What is the content of large text rolling or printing in the video?''} This consistent question formulation isolates the core challenge of extracting and integrating temporally distributed text while eliminating variability introduced by diverse question phrasings.
\end{itemize}

This standardized annotation approach for synthetic videos provides a controlled evaluation subset where performance differences can be attributed specifically to models' temporal text extraction.

\begin{table}[t]
\centering
\caption{Statistics of the ViTexQA dataset by domain category. \textbf{Avg. Duration:} Average video duration. \textbf{Avg. Q:} Average question length (number of words). \textbf{Avg. C+A:} Average CoT+Answer length (number of words).}
\label{tab:sta}
\resizebox{0.95\linewidth}{!}{
\begin{tabular}{lcccccc}
\hline
\textbf{Domain} & \textbf{\#Categories} & \textbf{\#Videos} & \textbf{\#Questions} & \textbf{Avg. Duration} & \textbf{Avg. Q} & \textbf{Avg. C+A} \\
\hline
Entertainment & 7 & 904 & 1,341 & 6.21 & 15.08 & 293.77 \\
Learning & 6 & 787 & 1,080 & 6.75 & 15.66 & 305.86 \\
Lifestyle & 6 & 743 & 940 & 6.71 & 17.26 & 316.37 \\
Business & 5 & 864 & 1,259 & 5.61 & 15.34 & 205.41 \\
Reality & 5 & 1,749 & 2,144 & 6.63 & 16.54 & 307.47 \\
Synthetic & 1 & 100 & 100 & 3.97 & 14.00 & 90.76 \\
\hline
 \rowcolor[HTML]{E5F4FB}
\textbf{TOTAL} & \textbf{30} & \textbf{5,147} & \textbf{6,864} & \textbf{6.32} & \textbf{16.00} & \textbf{253.27} \\
\hline
\end{tabular}
}
\end{table}

\subsection{Dataset Statistics and Analysis}

Upon completing the annotation process for both real and synthetic video instances, we present a comprehensive statistical overview of the resulting ViTexQA dataset. As shown in Table~\ref{tab:sta}, the dataset comprises \textbf{6,864} QA pairs drawn from \textbf{5,147} videos spanning \textbf{30} diverse domain categories, organized into five real-world domains (Entertainment, Learning, Lifestyle, Business, and Reality) and one synthetic domain, with an average video duration of \textbf{6.32} minutes. The average question length is \textbf{16.00} words and CoT+answer length is \textbf{253.27} words, substantially longer than M4-ViteVQA~\cite{zhao2022towards} (6.75 and 1.94 words, including only the answer), indicating richer semantic content. All queries are formulated as open-ended questions, with no multiple-choice or binary options included.
% 统计数据的表1

% We partition the ViTexQA dataset into training, validation, and test sets comprising 4,472 videos , 200 videos, and 475 videos, respectively. Unlike many existing video understanding benchmarks that only provide test sets with approximately 1,000 videos for zero-shot evaluation, ViTexQA includes a substantial training set to support both fine-tuning and zero-shot evaluation paradigms, which necessitates a smaller test set allocation. To ensure fair evaluation and prevent data leakage, we employ a stratified splitting strategy where each of the 30 scene categories is proportionally represented across all splits, and videos are strictly disjoint between splits with no overlap. This rigorous partitioning ensures that model performance on validation and test sets accurately reflects generalization capability to unseen videos rather than memorization of training data.

\section{Representative Examples from ViTexQA}
\begin{figure*}[h]
    \centering
    \includegraphics[width=0.95\linewidth]{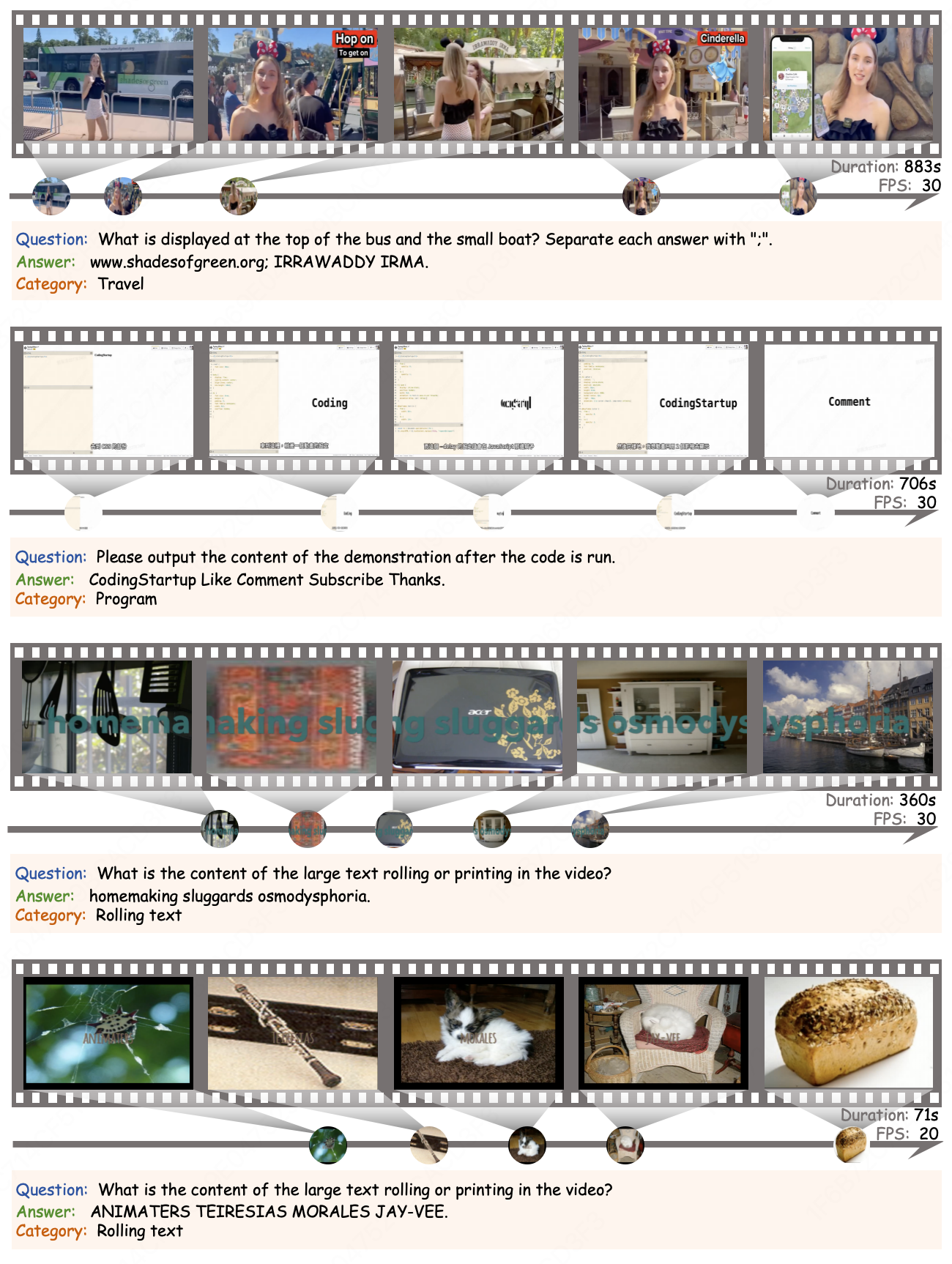}
    \caption{Representative Examples from ViTexQA.}
    \label{fig:example2}
\end{figure*}

To illustrate the diversity and temporal perception complexity inherent in ViTexQA, we present several representative examples that showcase the dataset's comprehensive coverage of real-world scenarios and multi-frame textual integration requirements, as shown in Fig.~\ref{fig:example2}. These examples demonstrate how questions necessitate cross-frame information synthesis, with answers emerging only through temporal aggregation of textual content distributed across multiple video frames, thereby highlighting the fundamental challenge that distinguishes ViTexQA from conventional single-frame text understanding tasks.

\section{Experiment Details}

\subsection{Training Details of CoT-Guided SFT}

The CoT-guided SFT stage of FrameThinker is built upon the Qwen3-VL-8B-Instruct base model and fine-tuned on the ViTexQA dataset. The model is trained in bfloat16 precision with Flash Attention 2~\cite{dao2023flashattention2} enabled for computational efficiency. We employ the AdamW optimizer for 1 epoch with a learning rate of $1 \times 10^{-4}$ and a warmup ratio of 0.05. To optimize memory utilization across GPUs, distributed training is conducted using DeepSpeed ZeRO Stage 3~\cite{rasley2020deepspeed}. The effective batch size per device is set to 2, achieved with 1 sample per device and 2 gradient accumulation steps.

\subsection{Training Details of Temporally-Grounded RL}

The Temporally-grounded RL stage builds upon the CoT-guided SFT checkpoint and further optimizes the model via Group Relative Policy Optimization (GRPO)~\cite{liu2024deepseek}, a reinforcement learning algorithm designed to improve response quality through relative reward signals. The model is initialized from the CoT-guided SFT checkpoint and further trained on the ViTexQA dataset. During GRPO training, we set the sampling temperature to 1.0, top-$p$ to 0.85, and repetition penalty to 1.1. For each training instance, the model generates 8 candidate completions with a maximum sequence length of 8,192 tokens, which are subsequently optimized using a composite reward function. We adopt the AdamW optimizer for 1 epoch with a learning rate of $1 \times 10^{-4}$ and a warmup ratio of 0.05. Training is conducted in bfloat16 mixed precision with gradient clipping at a maximum norm of 0.5 to ensure training stability. The GRPO algorithm applies a KL penalty coefficient $\beta = 0.001$ to balance reward maximization against deviation from the reference policy.

\subsection{Training Curves of FrameThinker}

\begin{figure*}[h]
    \centering
    \includegraphics[width=0.95\linewidth]{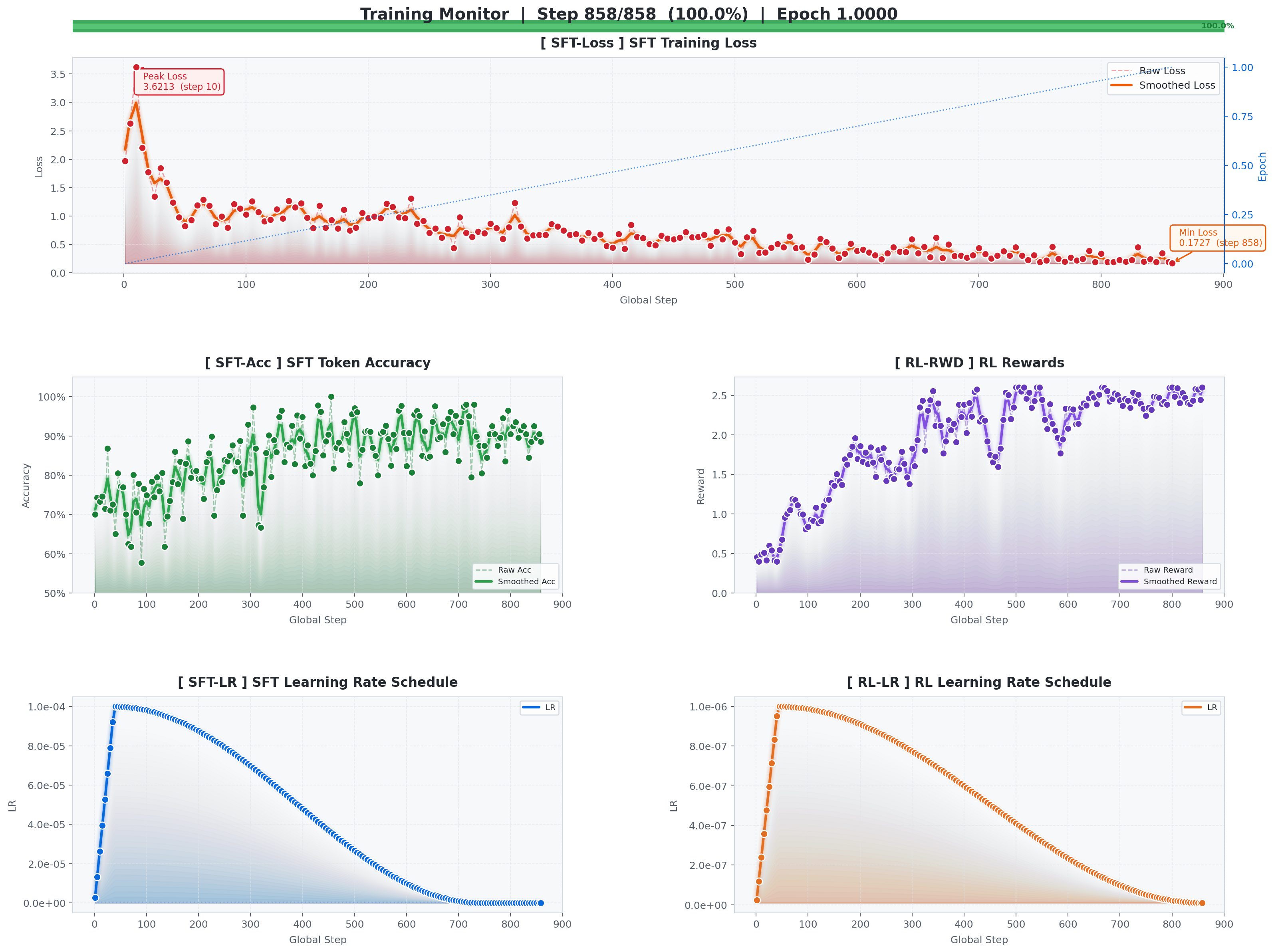}
    \caption{The training curves of our FrameThinker.}
    \label{fig:training_curves}
\end{figure*}

As illustrated in Figure~\ref{fig:training_curves}, we present five diagnostic plots that jointly characterize the training dynamics of FrameThinker across both the CoT-guided SFT and Temporally-grounded RL stages. Consistent and interpretable trends are observed throughout: training loss decreases steadily, while answer accuracy and reward scores exhibit continuous improvement, collectively demonstrating the stability and effectiveness of our two-stage training pipeline.

\begin{table*}[t]
\centering
\caption{Evaluation results of different metrics.}
\label{tab:metrics}
\scalebox{0.95}{ 
\begin{tabular}{
  >{\centering\arraybackslash}m{0.30\textwidth}
  >{\centering\arraybackslash}m{0.24\textwidth}
  >{\centering\arraybackslash}m{0.10\textwidth}
  >{\centering\arraybackslash}m{0.10\textwidth}
  >{\centering\arraybackslash}m{0.15\textwidth}
}\toprule
\textbf{Model Output} & \textbf{Ground Truth} & \textbf{Acc} & \textbf{ANLS} & \textbf{ROUGE-L} \\
\midrule
\rowcolor{green!06} 
\textbf{PLAYHOUSE SQUARE} & PLAYHOUSE SQUARE & 1.0 & 1.0 & 1.0 \\
 \rowcolor[HTML]{E5F4FB}
The content directly below the Earth pattern is \textbf{``internet''}, and the content below the black apple on the phone is \textbf{``iPhone''}. & internet; iPhone & 0.0 & 0.122 & 0.9375 \\
\rowcolor{green!06} 
The content above the image of a girl wearing VR glasses is \textbf{``Tech Learning with Professional Purpose''}. In the search box on the right phone, you can see the word \textbf{``View''} displayed. & Tech Learning with Professional Purpose; WWW & 0.0 & 0.2179 & 0.907 \\
 \rowcolor[HTML]{E5F4FB}
The yellow text content is \textbf{**Samsung M31s**}. & Rs. 14,999; Samsung M31s & 0.0 & 0.3095 & 0.6087 \\
\rowcolor{green!06} 
Jordyn; Madison; \textbf{Sydney;} \textbf{Gabe;} \textbf{Dominick;} \textbf{Lucas}. & JANELLE; MAX; SYDNEY; GABE; DOMINICK; LUCAS; JORDYN; MADISON; & 0.0 & 0.1967 & 0.2623 \\
 \rowcolor[HTML]{E5F4FB}
\textbf{Connections}; \textbf{to}; \textbf{Harvard}; CCP; members; attended & HARVARD AND CHINA; CONNECTIONS TO HARVARD & 0.0 & 0.0833 & 0.2 \\
\bottomrule
\end{tabular}}
\end{table*}

% \subsection{Model Configuration}

% For closed-source models including GPT-4o, Gemini-2.5 Pro, and Doubao-Seed-1.6, we adopt a uniform sampling rate of 1 fps with a maximum of 64 frames extracted per video, resized to 512 × 512 resolution to accommodate the input constraints of these models.

% For open-source models, we align each configuration with their original public implementations. LongVU, Qwen2.5-VL, and Qwen3-VL series operate under a 1 fps sampling strategy with a maximum of 768 frames extracted per video, resized to 448 × 448. Keye-VL adopts 1 fps sampling with a maximum of 768 frames at 336 × 336 resolution. InternVL3 applies a fixed 64-frame sampling window following its official implementation, with input resolution of 448 × 448. VideoLLaMA3 and LLaVA-Video are provided with extended temporal contexts—64 and 16 uniformly sampled frames respectively, resized to 336 × 336. MiniCPM-V4.5 and MiniCPM-V2.6 follow similar configurations, with 180 and 64 uniformly sampled frames respectively, resized to 448 × 448 and 448 × 448. 

\subsection{Evaluation metrics}

Since the questions in the ViTexQA dataset are formulated without explicit format constraints, certain models tend to generate extraneous content unrelated to the core answer, such as prefatory phrases including ``The characters in the video are$\cdots$,'' ``It looks like$\cdots$,'' or ``It seems that$\cdots$''. Additionally, answers are often embedded within quotation marks (`` ''), asterisks (* *), or positioned immediately following colons (:)$\cdots$. Even when using regularization to extract key content, other irrelevant information may still be extracted. To mitigate the interference of such irrelevant content and ensure robust evaluation of models' actual answer quality, we employ ROUGE-L scores as our primary evaluation metric to assess the output quality across different models.

ROUGE-L (Recall-Oriented Understudy for Gisting Evaluation - Longest Common Subsequence) recall is an evaluation metric based on the Longest Common Subsequence (LCS), designed to measure the coverage of model-generated text with respect to the ground truth.

Formally, let $\mathcal{R} = \{r_1, r_2, \ldots, r_m\}$ denote the reference text consisting of $m$ words, and let $\mathcal{C} = \{c_1, c_2, \ldots, c_n\}$ denote the candidate text comprising $n$ words, where $r_i$ and $c_j$ represent the $i$-th and $j$-th word token in the respective sequences. The ROUGE-L score is then defined as:

\begin{equation}
s_{\mathrm{lcs}} = \frac{\mathrm{LCS}(\mathcal{R}, \mathcal{C})}{m}
\end{equation}
where $\mathrm{LCS}(\cdot, \cdot)$ represents the length of the longest common subsequence between reference text $\mathcal{R}$ and candidate text $\mathcal{C}$, and $m$ is the total number of words in the reference text.

To intuitively compare the performance of commonly used metrics, we present a comparison of Acc (Accuracy), ANLS (Average Normalized Levenshtein Similarity), and ROUGE-L scores between the outputs of several models and the ground truth in Table~\ref{tab:metrics}. As shown in the table, Acc fails to capture partial correctness and assigns a score of 0.0 to all imperfect outputs regardless of their actual overlap with the ground truth. ANLS is sensitive to character-level edit distance, which makes it susceptible to minor formatting differences such as punctuation and capitalization, leading to disproportionately low scores even when the semantic content is largely correct. In contrast, ROUGE-L demonstrates greater robustness by rewarding subsequence-level matches, effectively capturing the core answer content while tolerating extraneous surrounding text, making it the most suitable metric for evaluating open-ended video text understanding tasks.

\section{Broader Impact}

The introduction of ViTexQA and FrameThinker advances the development of temporal perception multimodal AI systems capable of genuine cross-frame textual perception. By establishing a rigorously verified benchmark where 100\% of questions require multi-frame integration, our work addresses a critical evaluation gap that has allowed prior models to circumvent temporal perception through single-frame shortcuts. This contribution complements recent efforts in video understanding~\cite{wang2024internvideo2,shen2025long,liu2025nvila} and multimodal perception~\cite{bai2025qwen2,chen2024internvl}, providing a more faithful measure of progress toward human-level video comprehension.

\noindent\textbf{Positive Societal Impact.}
Improved video text understanding has broad practical relevance. In accessibility contexts, systems capable of reliably extracting and perception over temporally distributed captions can benefit users with hearing impairments by enabling more accurate automatic summarization of captioned broadcasts~\cite{jahagirdar2023watching}. In education and information retrieval, models trained on ViTexQA can better support comprehension of instructional videos and news content where critical information unfolds progressively across frames~\cite{lee2025video,de2025describe}. More broadly, our open benchmark and human-centric annotation methodology provide the research community with reusable resources for developing and evaluating temporally grounded video understanding systems.

\noindent\textbf{Limitations and Potential Risks.}
We acknowledge several limitations that are inherent to the nature of large-scale video data collection and are largely beyond the control of individual research efforts. First, although ViTexQA spans 30 diverse domain categories, the source videos are predominantly drawn from publicly available platforms (e.g., YouTube) and existing benchmark datasets (with original annotations discarded). The demographic, linguistic, and cultural distributions of such content are determined by the broader ecosystem of online video production and platform recommendation mechanisms—factors that lie outside the direct control of dataset curators. As a result, models trained or evaluated exclusively on ViTexQA may exhibit degraded performance on underrepresented populations, non-English textual content, or culturally specific visual contexts. Importantly, this limitation is not a reflection of methodological shortcomings but rather an unavoidable characteristic shared by virtually all large-scale web-sourced benchmarks~\cite{yin2024survey,alampara2025probing}. In contrast, our core contribution—the rigorous human-centric annotation pipeline, the temporally grounded QA construction methodology, and the FrameThinker training framework—remains fully valid and generalizable, as these components are independent of the specific video source distribution. Second, as with any system that improves video text understanding, there exists a potential risk of misuse in surveillance or privacy-invasive applications. We emphasize that ViTexQA is strictly intended for academic research purposes, and all collected data complies with applicable data privacy policies and platform terms of service. We further encourage the community to develop responsible deployment guidelines when extending models trained on ViTexQA to real-world applications.

\noindent\textbf{Recommendations for Future Work.}
To mitigate the above concerns, we encourage future work to:
\begin{itemize}
    \item Expand the dataset to include multilingual and culturally diverse video content.
    \item Investigate fairness-aware evaluation protocols that assess model performance across demographic subgroups.
    \item Develop deployment guidelines that accompany the release of models trained on ViTexQA.
\end{itemize}
We believe that proactive engagement with these issues is essential for ensuring that advances in video text understanding translate into equitable and beneficial real-world applications.

\begin{figure*}[h]
    \centering
    \includegraphics[width=0.95\linewidth]{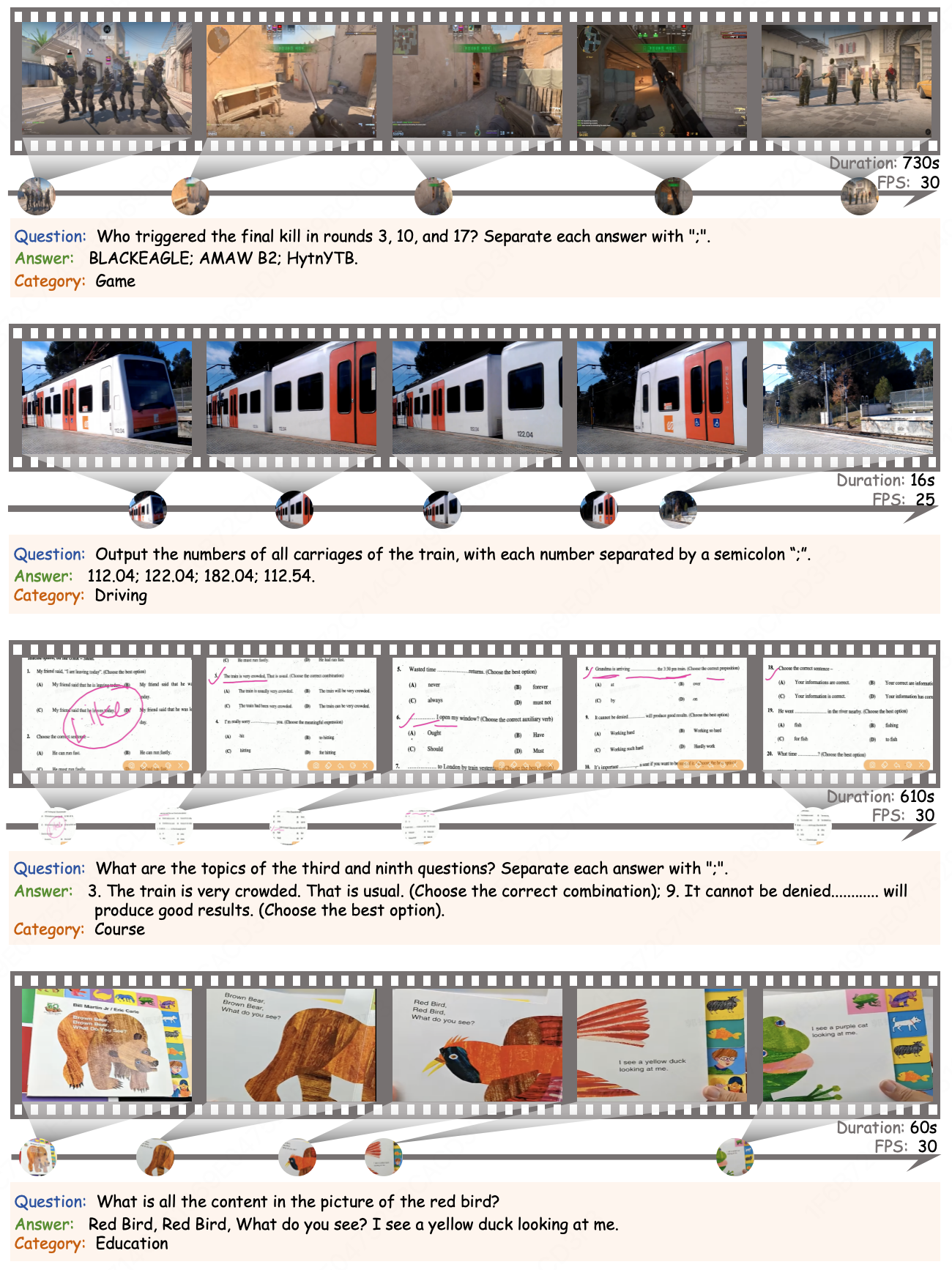}
    % \caption{Overview  }
    \label{fig:example1}
\end{figure*}

% WARNING: do not forget to delete the supplementary pages from your submission 

% \clearpage  % TODO FINAL: This \clearpage needs to be removed from both review and camera-ready versions.

% \section*{Acknowledgements}
% Please insert your acknowledgments here.

% ---- Bibliography ----
%
% BibTeX users should specify bibliography style 'splncs04'.
% References will then be sorted and formatted in the correct style.
%
% \bibliographystyle{splncs04}
% \bibliography{main}
\end{document}